\documentclass[10pt,twocolumn,letterpaper]{article}

\usepackage[pagenumbers]{cvpr} 

\definecolor{cvprblue}{rgb}{0.21,0.49,0.74}
\usepackage[pagebackref,breaklinks,colorlinks,allcolors=cvprblue]{hyperref}

\usepackage{graphicx}
\usepackage{amsmath}
\usepackage{amssymb}
\usepackage{booktabs}
\usepackage{microtype}

\usepackage{float}
\usepackage{caption}
\usepackage{subcaption}
\usepackage[all]{nowidow}

\usepackage{makecell}
\usepackage{placeins}
\usepackage{tabularx}

\usepackage[capitalize]{cleveref}
\crefname{section}{Sec.}{Secs.}
\Crefname{section}{Section}{Sections}
\Crefname{table}{Table}{Tables}
\crefname{table}{Tab.}{Tabs.}

\usepackage{etoolbox}
\newcounter{BalanceAtReference}
\newcounter{ReferenceIndexForBalancing}
\makeatletter
\def\@balancelastpageonce{%
	\ifnum\value{ReferenceIndexForBalancing}=\value{BalanceAtReference}
	\newpage
	\else
	\relax
	\fi
	\stepcounter{ReferenceIndexForBalancing}
}
\pretocmd{\bibitem}{\@balancelastpageonce}
{} 
{\@latex@error{Patching \bibitem failed}{\@ehd}}
\makeatother

\def\paperID{21240} 
\def\confName{CVPR}
\def\confYear{2026}

\begin{document}

\title{Scriboora: Rethinking Human Pose Forecasting}

\author{Daniel Bermuth\\
University of Augsburg, Germany\\
{\tt\small daniel.bermuth@uni-a.de}
\and
Alexander Poeppel\\
University of Augsburg\\
{\tt\small poeppel@isse.de}
\and
Wolfgang Reif\\
University of Augsburg\\
{\tt\small reif@isse.de}
}

\maketitle


\begin{abstract}
   Human pose forecasting predicts future human poses from past observations for applications including action recognition, sports analysis, and human-robot interaction.
   However, progress in the field is difficult to assess because reported results often rely on heterogeneous preprocessing, inconsistent metric implementations, and partially released artifacts.
   This paper re-examines human pose forecasting as an evaluation problem by auditing a wide range of recent forecasting methods, identifying several reproducibility issues, and introducing a unified training and evaluation pipeline for absolute pose forecasting.
   Under this unified protocol, mature cross-domain sequence models adapted from speech recognition provide strong baselines and obtain the lowest errors among the evaluated models.
   To study deployment-relevant noise, a paired clean/detector-noisy dataset variant is introduced using poses generated by a pose-estimation model.
   The experiments show that detector-generated poses cause substantial performance degradation compared with clean motion-capture inputs, but that part of this degradation can be recovered by adapting the forecaster on detector-generated pose sequences.
\end{abstract}


\section{Introduction}

\begin{figure}[tp!]
   \centering
   \includegraphics[width=0.77\linewidth]{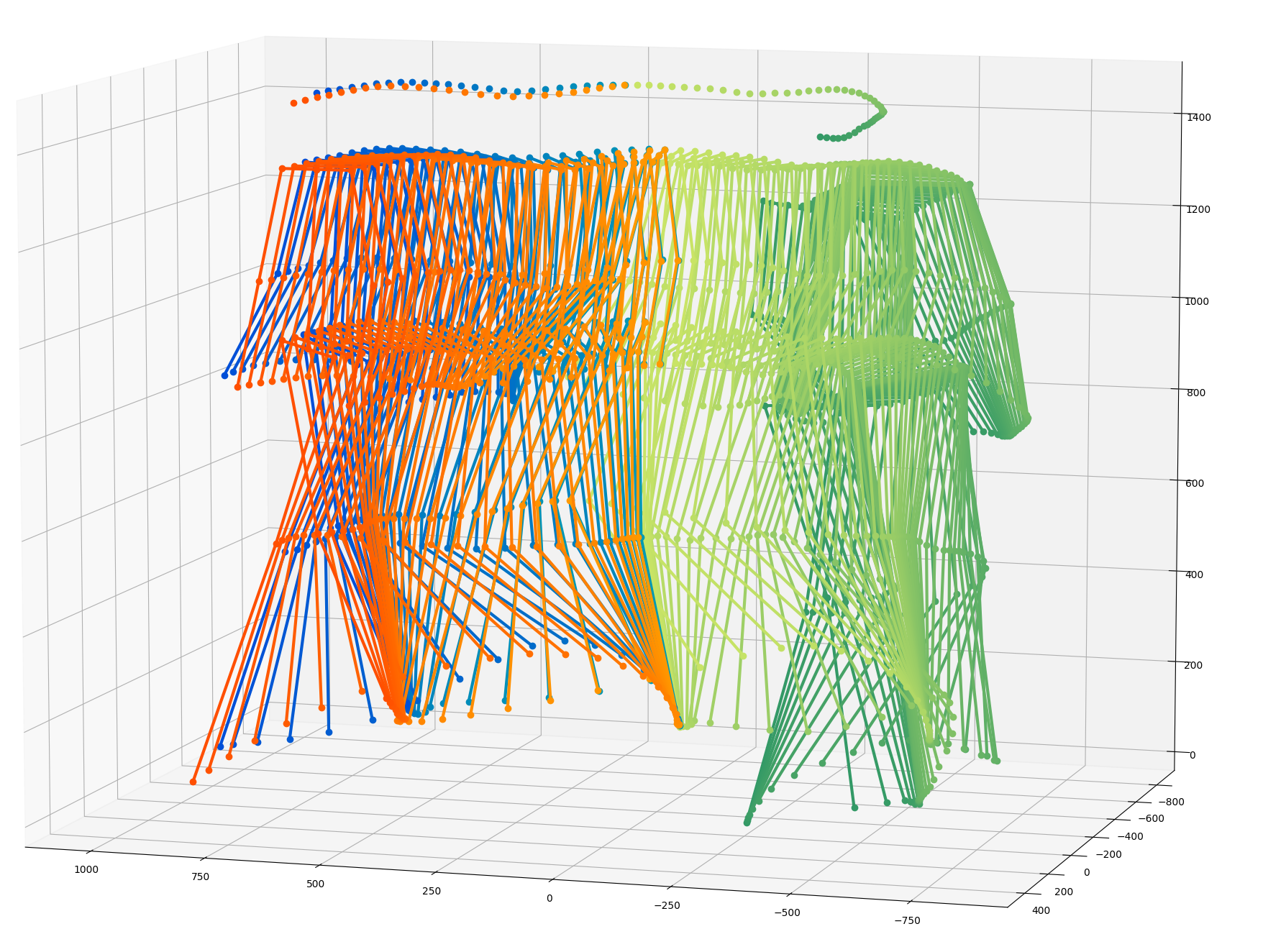}
   \caption{\looseness=-1 Example of an \textit{absolute} pose forecast of a walking person. The green skeletons visualize the input sequence of the prediction model, the red ones the predicted future poses, and the blue ones the ground-truth labels.}
   \label{fig:init_example}
\end{figure}

\looseness=-1
Human pose forecasting represents observed motion as sequences of body-joint coordinates and predicts future joint trajectories from past observations (see Figure~\ref{fig:init_example}).
Such forecasts can support applications such as predicting pedestrian movements or replanning robot motions in collaborative settings.

Although many methods report rapid progress, the state of the field remains difficult to assess.
Results are often produced with heterogeneous preprocessing, inconsistent metric implementations, partially released artifacts, and sometimes evaluation errors.
Comparisons across papers are therefore unreliable, and claimed improvements can vanish once consistent protocols are applied.
Prior evaluations also tend to compare only against specialized pose-forecasting baselines, leaving unclear whether mature sequence models from neighboring domains already provide strong baselines.
At the same time, evaluations rarely reflect deployment conditions.
In practical systems, input poses often come from markerless human pose estimators, since outfitting pedestrians or factory workers with tracking suits is usually infeasible.
The errors of such estimators differ from synthetic noise used in prior studies and remain underexplored for pose forecasting.

\looseness=-1
This paper addresses these issues with a practice-oriented evaluation study of absolute pose forecasting, making three contributions:
(1)~a broad reproducibility audit of recent forecasting methods;
(2)~a unified absolute-pose evaluation showing that prior rankings do not necessarily transfer and that mature cross-domain sequence baselines obtain lower errors than the evaluated specialized models under the same protocol;
(3)~a paired clean/detector-noisy evaluation for measuring pose-estimator-induced degradation, together with pseudo-label adaptation to recover part of the lost performance.
Taken together, these contributions are intended to provide a repeatable and application-targeted benchmark protocol that enables future research to better compare and evaluate pose-forecasting methods.

To support subsequent work in these directions, model and preprocessing code, evaluation tools, and proposed-model checkpoints are released at:~\url{https://gitlab.com/Percipiote/}

\section{Related Work}

Most recent pose forecasting approaches have specialized into two categories:
(1) \textit{Relative} pose forecasting, which predicts a future pose relative to the last input pose, and therefore does not include global human motion trajectories~(\textit{HisRepItself}~\cite{wei2020his}, \textit{STSGCN}~\cite{sofianos2021spacetimeseparable}, \textit{SPGSN}~\cite{li2022skeleton}, \textit{PGBIG}~\cite{ma2022progressively}, \textit{MotionMixer}~\cite{ijcai2022p111}, \textit{siMLPe}~\cite{guo2022back}, \textit{EqMotion}~\cite{xu2023eqmotion}, \mbox{\textit{DePOSit}}~\cite{saadatnejad2023generic}, \textit{AuxFormer}~\cite{xu2023auxiliary}, \textit{GMFnet}~\cite{shi2024gradient}, \textit{GCNext}~\cite{wang2024gcnext}, \textit{DeformMLP}~\cite{huang2024deformmlp}).
(2) \textit{Absolute} pose forecasting, which predicts a future pose in absolute coordinates, and therefore also includes global human motion trajectories~(\textit{SoMoFormer}~\cite{vendrow2022somoformer}, \mbox{\textit{TBiFormer}}~\cite{peng2023trajectory}, \textit{JRTransformer}~\cite{xu2023joint}, \textit{IAFormer}~\cite{xiao2024multi}, \textit{EMPMP}~\cite{zheng2025efficient}).
Appendix~\ref{app:relwork} provides additional architectural details.

The \textit{mean per joint position error} (\textit{MPJPE}, Equation~\ref{eq:mpjpe}) is the most commonly used metric to compare different approaches.
It computes the distance between each predicted joint position and the ground-truth position.
Afterward, those distances are averaged across all joints and persons to obtain a single score.
When considering a forecast with multiple timesteps, the \textit{MPJPE} score is only calculated for the prediction of a specific timestep.
\textit{MPJPE} is reported in millimeters throughout this paper.

\vspace{-6pt}
\begin{equation}
   \small
   MPJPE_{t} = \frac{ \sum\limits_{\substack{p=1\\j=1}}^{\substack{persons\\joints}} \sqrt{\sum\limits_{k=1}^{3} (gt_{t,p,j,k} - pred_{t,p,j,k})^2}}{persons \cdot joints}
   \label{eq:mpjpe}
\end{equation}

Depending on the two different categories, the commonly used datasets and evaluation protocols differ as well.
For relative pose forecasting, the standard dataset is the \textit{Human3.6M}~\cite{h36m_pami} dataset, where actors are recorded by 4 cameras placed in the room corners while performing 15~different actions.
Most relative approaches build upon the source-codes from~\cite{martinez2017simple, martinez2017human} and reuse their preprocessed datasets and evaluation metrics.
The absolute pose forecasting approaches, on the other hand, are usually evaluated on the \textit{3D Poses in the wild}~(3DPW) dataset~\cite{von2018recovering}, which consists of 60~action sequences in a real-world setting, or the \textit{CMU-MoCap}~\cite{mocap} dataset, in which the actors have been captured by twelve cameras in a $3m\,$$\times$$\,8m$ room.

Most of those approaches have focused on using ground-truth pose labels from the dataset, and only a few approaches~\cite{cui2021towards, saadatnejad2023generic} also added artificial errors.
For several reasons, which are explained in more detail in Section~\ref{chap:noisy}, such artificial errors differ from the pose estimation errors.
Therefore, the impact of such errors will be investigated later on, using paired clean and detector-generated pose sequences, which allow the detector-induced forecasting error to be measured against ground truth.


\section{Reproducibility Audit}
\label{sec:reprod}

Before constructing a unified benchmark, the reported results of recent relative and absolute pose-forecasting methods were audited using the authors' released artifacts where available.
The goal of the audit is not only to check whether numbers can be rerun, but also to identify which evaluation assumptions make reported comparisons fragile.

\vspace{3pt}
For each method, the audit recorded whether source code, checkpoints, preprocessing scripts, and evaluation scripts were available.
It then tested whether the reported numbers could be reproduced without modifying the released evaluation procedure (allowing differences of up to $1mm$).
When reproduction was unsuccessful, the dominant discrepancy source was categorized as a metric implementation issue, missing or partial artifacts, checkpoint-only reproducibility, training mismatch, unit or scale ambiguity, timestamp ambiguity, or environment failure.
When a released evaluation procedure contained an identifiable metric bug, both the original reproduced behavior and the corrected result were recorded.
Table~\ref{tab:audit_summary} summarizes the observed failure modes, while detailed numerical comparisons and per-method notes are provided in Appendix~\ref{app:audit_details}.

\begin{table}[htpb!]
\centering
{
\setlength{\tabcolsep}{3pt}
\fontsize{7.2}{9.5}\selectfont
\begin{tabularx}{\columnwidth}{llX}
   \toprule
   Method & Reproduction outcome & Main issue \\
   \midrule

   MotionMixer~\cite{ijcai2022p111}& corrected result lower & averaging/counting bug  \\

   STSGCN~\cite{sofianos2021spacetimeseparable} & corrected result lower & MPJPE implementation bug  \\

   GMFnet~\cite{shi2024gradient} & checkpoint reproduced & training/checkpoint mismatch  \\

   SPGSN~\cite{li2022skeleton} & retrained result lower & unresolved training mismatch  \\

   HisRepItself~\cite{wei2020his} & reproduced by code & -- \\

   AuxFormer~\cite{xu2023auxiliary} & checkpoint reproduced & training/checkpoint mismatch \\

   PGBIG~\cite{ma2022progressively} & reproduced by code & --   \\

   EqMotion~\cite{xu2023eqmotion} & checkpoint reproduced & training/checkpoint mismatch \\

   GCNext~\cite{wang2024gcnext} & reproduced by code & --  \\

   CSCNet~\cite{li2025component} & no code to reproduce & missing artifact  \\

   DePOSit~\cite{saadatnejad2023generic} & checkpoint reproduced & training mismatch unresolved \\

   siMLPe~\cite{guo2022back} & reproduced by code & --   \\

   DeformMLP~\cite{huang2024deformmlp} & reproduced by code & --   \\

   IAFormer~\cite{xiao2024multi} & partially reproduced & missing H36M implementation  \\

   TRiPOD~\cite{adeli2021tripod} & metric discrepancy & partial code, unintended metric behavior  \\

   MRT~\cite{wang2021multi}  & ambiguous & partial code, metric unit ambiguity  \\

   SoMoFormer~\cite{vendrow2022somoformer}  &  ambiguous & partial code, metric unit ambiguity  \\

   TBiFormer~\cite{peng2023trajectory} &ambiguous & partial code, metric unit ambiguity  \\

   T2P~\cite{jeong2024multi} & not evaluated & partial code, environment setup failure  \\

   EMPMP~\cite{zheng2025efficient} &   ambiguous & partial code, metric unit ambiguity  \\

   JRTransformer~\cite{xu2023joint} &   ambiguous & partial code, metric unit ambiguity  \\

   \bottomrule
\end{tabularx}
}
\vspace{6pt}
\caption{Summary of reproducibility audit. The table reports the dominant issue observed for each method. ``Checkpoint reproduced'' means that the released checkpoint matched the reported result, but training from the released code did not reproduce the same performance.  ``Reproduced by code'' means that the model was successfully trained and evaluated using the publicly released code. ``Corrected result lower'' means that an evaluation issue was identified and the corrected score was worse than the originally reported score. Detailed numerical comparisons and forensic notes are provided in Appendix~\ref{app:audit_details}.}
\label{tab:audit_summary}
\end{table}

The audit showed that reproducibility problems are not limited to missing code.
Several discrepancies arised from evaluation design choices that directly affect scientific conclusions, including metric implementations, preprocessing conventions, unit conversions, and forecast timestamps.
In particular, correcting metric errors changed the apparent ranking of some methods, while incomplete preprocessing releases and ambiguous scale conventions made several absolute-pose benchmarks difficult to compare across papers.
Overall, substantial reproducibility or evaluation issues were found across many of the examined methods.
Reported improvements therefore cannot be interpreted reliably unless preprocessing, metric implementation, units, timestamps, and released artifacts are controlled.
These findings motivate the unified protocol in Section~\ref{sec:unified} and highlight the need to release complete preprocessing and evaluation code together with trained checkpoints where possible.


\FloatBarrier
\section{Latency-Aware Metrics}
\label{sec:metrics}

Many uses of pose forecasting are latency-sensitive, but standard forecasting metrics ignore the time required to compute a forecast (measured by \textit{forecasts per second (FPS)} at batch-size $1$).
For example, a model that requires more than one second to forecast the next second of motion has limited value in a real-time system, because the predicted interval has elapsed before the forecast is available.
To measure how inference time affects the usefulness of a forecast once pose inputs are already available, two new latency-aware metrics are introduced.
These metrics are intended to complement \textit{MPJPE} when comparing forecasting models under a fixed upstream sensing pipeline.

The \textit{forecast after delay error} (\textit{FADE}, Equation~\ref{eq:fade}) focuses on average forecasting performance while accounting for inference delay.
The idea is that if a prediction takes some amount of time to compute, the forecast must effectively reach that much further into the future.
Because changing the output sequence length of every model would be impractical, \textit{FADE} uses a simpler approximation that assumes linear error growth with time.
Appendix~\ref{app:fade} validates the linearity assumption in more detail.

\begin{equation}
   \small
   FADE_{t} = MPJPE_{t} + MPJPE_{t} \cdot \frac{1000\,ms}{t}  \cdot  \frac{1}{FPS}
   \label{eq:fade}
\end{equation}

The \textit{fast change estimate} (\textit{FCE}, Equation~\ref{eq:fce}) is a conservative forecaster-side latency measure for abrupt motion changes.
Using the human limb speed from \textit{ISO\,13855}, it estimates how far a limb could move during one model inference interval.
After an unexpected motion change, the system relies on a prediction conditioned on outdated observations until a new forecast is computed.
Lower \textit{FCE} indicates faster reaction to sudden motion changes.

\begin{equation}
   \vspace{-3pt}
   \small
   FCE = 2000\,mm \cdot  \frac{1}{FPS}
   \label{eq:fce}
   \vspace{3pt}
\end{equation}

In real applications, the image-to-pose system also introduces latency.
This latency is excluded from \textit{FADE} and \textit{FCE} because it is deployment-specific and depends on the camera and hardware systems as well as on the pose-estimation method.
For a fixed upstream pipeline, this latency is shared across forecasters and therefore does not affect their relative comparison.
End-to-end deployment evaluations should add upstream sensing and pose-estimation latency separately.


\section{Unified Absolute-Pose Benchmark}
\label{sec:unified}

Forecasting absolute human poses can be important for a variety of applications such as motion planning, virtual reality or human-robot interaction. 
If a person moves around, their absolute position in the world can contain important information that is not captured by relative pose forecasts, for example future distances between human joints and objects or devices in the environment, and therefore would require an additional trajectory prediction model.

Rather than building a more complex two-model system that stacks trajectory and relative forecasting models, and then also their prediction errors, this work evaluates whether the already existing relative forecasting algorithms can be adapted to directly predict absolute poses as well.

To use the relative algorithms with absolute poses, the input pipeline was replaced.
Instead of reusing the preprocessed data without global motion, which all approaches took from~\cite{martinez2017human}, the absolute poses were extracted directly from the \textit{Human3.6M} dataset. This avoids potential sources of error introduced by repeated conversions between absolute and relative pose representations and additionally simplifies the overall data processing pipeline.

To improve the training generalization, the input and target pose sequences were moved to a central location by subtracting the mid-hip joint of the last input pose from all joint coordinates.
For inference, the already known last input hip position is added again, to obtain global coordinates.
The subtraction ensures that the model learns location-independent motion patterns, even though it might be interesting in some usecases to keep this information, for example, if movements only happen at certain locations.

For the main comparison, the \textit{Human3.6M}~\cite{h36m_pami} dataset is used, downsampled to $25Hz$ by skipping every second step, as in previous works.
The models are trained on $50$ input and $25$ output timesteps, resulting in~$1s$~of motion forecast, and only use 13~joints (2~hips, 2~shoulders, 1~nose, 2~knees, 2~ankles, 2~elbows, 2~wrists) to simplify usage with other datasets that have fewer ground-truth joint labels.
The MPJPE is calculated action-agnostically by averaging over all dataset samples.
All models are trained from scratch and evaluated on the same data, using released default hyperparameters whenever possible (required compatibility modifications are documented in Appendix~\ref{app:convert}, and code and configuration files are included in the open-source repository).

\begin{table}[htbp]
  \centering
   {
   \setlength{\tabcolsep}{4pt}
   \fontsize{8}{9}\selectfont
   \begin{tabular}{@{}|l|c|c|c|r|r|c|@{}}
      \toprule
      Method                              & Type & Size & MPJPE & FPS & \hspace{-1pt}FCE\hspace{-1pt} & FADE \\
      \midrule
      \textit{Repeat last frame}        & -    & -     & 301 &  - { } & - { } &  301 \\
      \textit{Last delta average}       & -    & -     & 264 &  - { } & - { } &  264 \\
      \textit{Ridge Regression}         & -    & -     & 185 &  - { } & - { } &  185 \\
      TBiFormer      \hspace{10pt}'2023 & abs & 5.64M & 301 & 293    & 7     &  302 \\
      PGBIG          \hspace{22pt}'2022 & rel & 2.65M & 279 & 35     & 57    &  287 \\
      GMFnet         \hspace{19pt}'2024 & rel & 11.3M & 279 & 51     & 39    &  284 \\
      SPGSN          \hspace{21pt}'2022 & rel & 5.21M & 262 & 19     & 105   &  276 \\
      STSGCN         \hspace{15pt}'2021 & rel & 202K  & 275 & 1405   & 1     &  275 \\
      HisRepItself   \hspace{05pt}'2020 & rel & 2.93M & 233 & 142    & 14    &  235 \\
      DePOSit        \hspace{17pt}'2023 & rel & 1.17M & 197 & 6      & 333   &  230 \\
      GCNext         \hspace{19pt}'2024 & rel & 4.95M & 175 & 38     & 53    &  180 \\
      siMLPe         \hspace{21pt}'2022 & rel & 129K  & 178 & 158    & 13    &  179 \\
      SoMoFormer     \hspace{03pt}'2022 & abs & 4.88M & 179 & 586    & 3     &  179 \\
      MotionMixer    \hspace{03pt}'2022 & rel & 37.2K & 168 & 632    & 3     &  168 \\
      IAFormer       \hspace{14pt}'2024 & abs & 2.53M & 158 & 28     & 71    &  164 \\
      AuxFormer      \hspace{09pt}'2023 & rel & 1.27M & 164 & 403    & 5     &  164 \\
      JRTransformer  \hspace{-1pt}'2023 & abs & 3.74M & 162 & 595    & 3     &  162 \\
      DeformMLP      \hspace{05pt}'2024 & rel & 1.29M & 159 & 512    & 4     &  159 \\
      EMPMP          \hspace{18pt}'2025 & abs & 314K  & 150 & 215    & 9     &  151 \\
      EqMotion       \hspace{14pt}'2023 & rel & 3.36M & 148 & 45     & 44    &  151 \\
      \bottomrule
   \end{tabular}
   }
  \vspace{6pt}
  \caption[Unified comparison of \textit{absolute} pose forecasting on \textit{Human3.6M}]{Unified comparison of \textit{absolute} and \textit{relative} methods in the task of \textit{absolute} pose forecasting on \textit{Human3.6M}. \textit{MPJPE} and \textit{FADE} are measured at timestep $1000\,ms$. The error metrics are in millimeters. \textit{FPS}~is tested on a computer with an AMD-9900X and a single Nvidia-RTX4080. A batch-size of~$1$ was used for testing. Note that the size and speed of some models depends on the number of input and output timesteps (here: $50$ in, $25$ out).}
   \label{tab:abs_h36m}
\end{table}

\FloatBarrier
Table~\ref{tab:abs_h36m} shows that several relative forecasting models can be adapted successfully to absolute pose forecasting.
A new static approach was added for comparison as well, \textit{last delta average} repeats the average of the last input pose delta, so all joints keep moving in the direction of the last general movement (the average delta is used to preserve the human shape over the time).
As an additional classic method baseline, a simple \textit{Ridge Regressor} was trained.

\looseness=-1
An interesting finding is that most models using graph neural networks (\textit{HisRepItself}, \textit{STSGCN}, \textit{SPGSN}, \textit{PGBIG}, \textit{GMFnet}) perform worse in this unified setting than the others using a different network structure.
This suggests that conclusions drawn from the original protocols may not directly transfer to this absolute-forecasting protocol, but it should not be read as evidence that graph-based models are generally inferior.
At the same time, the results show that the general idea of switching the task of the relative models was successful, because some of those adapted implementations obtain lower errors than several evaluated absolute-pose implementations, including \textit{SoMoFormer}, \textit{JRTransformer}, and \textit{EMPMP}.


\section{Cross-Domain Sequence Baselines}
\label{sec:cdsb}

The results in Table~\ref{tab:abs_h36m} show that diverse architectures obtain similar results and that several relative-pose models transfer to the absolute-pose protocol.
From this observation, it remains unclear to what extent pose-forecasting performance depends on specialized architectures.
This motivated testing whether mature generic sequence architectures, originally introduced years earlier, can match or outperform recent pose-specific models after minimal adaptation.
The goal was not to identify the best possible architecture, but to test whether current baseline comparisons are incomplete.

\subsection{Scriboora}

\looseness=-1
The idea behind \textit{Scriboora} is the question: \textit{Can speech-to-text models be retasked as well?}
At a high level, both speech recognition and pose forecasting transform an observed numerical sequence into a target sequence, although the target spaces differ and are either character-tokens or joint coordinates.
Furthermore, speech recognition provides many mature sequence models with diverse architectures that emphasize efficient inference while capturing local temporal patterns and long-range sequence structure, all of which are relevant to pose forecasting.
For the experiments with different speech-to-text models (\textit{DeepSpeech}~\cite{DEPSPE}, \textit{QuartzNet}~\cite{kriman2020quartznet}, \textit{Conformer}~\cite{gulati2020conformer}, \textit{Squeezeformer}~\cite{kim2022squeezeformer}), the implementations from \textit{Scribosermo}~\cite{scribosermo} were used, which also motivated the first part of the paper title, \textit{Scriboora}.

\subsection{Performance Results}

The main change needed was to adapt the data pipeline, so that motion instead of audio sequences could be loaded. 
The models themselves only required minor changes.
Appendix~\ref{app:scriboora} provides details on the model changes and training procedure.
Table~\ref{tab:abs_stt} shows that the converted speech models are strong baselines under the unified protocol, with several matching or surpassing specialized models from Table~\ref{tab:abs_h36m}.
The \textit{Conformer} models combine convolutional, linear, and attention layers to capture both local and global patterns and are therefore among the leading approaches in speech recognition. This relatively generic architecture also appears to adapt well to the new task.

\begin{table}[H]
  \centering
  \fontsize{8pt}{9pt}\selectfont
  \begin{tabular}{@{}|l|c|c|r|r|c|@{}}
    \toprule
    Method                                 & Size  & MPJPE & FPS  & \hspace{-1pt}FCE\hspace{-1pt}  & FADE \\
    \midrule
    DeepSpeech                      \hspace{07pt}'2014 & 2.35M & 198   & 653  &  3 &  198 \\
    QuartzNet                       \hspace{14pt}'2020 & 3.17M & 168   & 1300 &  2 &  168 \\
    Squeezeformer                   \hspace{-1pt}'2022 & 3.31M & 149   & 960  &  2 &  149 \\
    Conformer                       \hspace{12pt}'2020 & 3.22M & 147   & 1049 &  2 &  147 \\
    \midrule
    MotionConformer                          & 9.23M & \textbf{143}   & 929  &  2 &  \textbf{143} \\
    \bottomrule
  \end{tabular}
  \vspace{6pt}
  \caption[Results of converted speech models on \textit{Human3.6M}]{Results of converted speech models on \textit{Human3.6M}.}
  \label{tab:abs_stt}
\end{table}

\looseness=-1
As can be seen in Table~\ref{tab:abs_stt} as well, most models are very fast, so this leaves room to further increase the size of future models.
With that in mind, the \textit{Conformer} model was further improved by increasing the layer dimensions, and as another change, because the motion sequence length is notably shorter than the average audio sequence length, the time downsampling at the start of the model was moved to the end of the model, so that information is retained longer in the model (see Appendix~\ref{app:scriboora}).
The resulting model is called \textit{MotionConformer} in the following.
Although it is the second-largest model overall, the training time of $1h$ on the Nvidia-RTX4080 is at the lower end among the compared architectures.
As shown in Table~\ref{tab:abs_stt}, it obtains the lowest reported error among the models evaluated under the unified protocol, while maintaining a very high inference speed as well.


\section{Further Forecast Benchmarks}
\label{app:more_benchs}

The previous benchmark evaluated a broad set of methods on the unified \textit{Human3.6M} protocol.
To test whether the main trends persist beyond this dataset, this section reports additional experiments on \textit{CMU-MoCap} and \textit{CHI3D}.
Because retraining every audited model on every additional dataset would be computationally expensive, these experiments focus on the strongest methods from the unified benchmark and on representative baselines.

\subsection{Single-person Forecasts}
 
For the single-person evaluation, the \textit{CMU-MoCap}~\cite{mocap} dataset was selected as an additional absolute-pose forecasting benchmark.
The labels were extracted from the original dataset again, and are different to the ones used in \textit{MRT}, \textit{SoMoFormer}, \textit{JRTransformer} and \textit{EMPMP}.
The processing script of \textit{MRT} adds a randomly chosen global position offset, and also combines single actor sequences with other sequences to get sequences containing three actors, which are then cut to fixed sized $4s$ sub-sequences.
Since for this experiment only single-actor sequences are evaluated, sequences with two actors are split into two. 
The performance of the models trained and tested on \textit{CMU-MoCap} can be found in Table~\ref{tab:abs_mocap1}.
The relative performance between the models is similar to the one on \textit{Human3.6M}, with \textit{MotionConformer} again obtaining the lowest errors.

\begin{table}[H]
   \centering
   \fontsize{8pt}{9pt}\selectfont
   \begin{tabular}{|l|c|c|r|c|}
      \toprule
      Method & Size & \makecell{MPJPE \\ 400 / 1000} & FPS & \makecell{FADE \\ 400 / 1000} \\
      \midrule
      \textit{Repeat last frame}           & - & { } 201 / 408                    & - { } & 201 / 408   \\
      \textit{Last delta average}          & - & { } 167 / 398                    & - { } & 167 / 398   \\
      \textit{Ridge Regression}            & - & { } 130 / 296                    & - { } & 130 / 296   \\
      JRTransformer                        & 3.79M & 99.0 / 245                         & 610   & 99.4 / 245  \\
      DeformMLP                            & 1.29M & 98.3 / 242                         & 462   & 98.8 / 243  \\
      EqMotion                             & 3.37M & 90.6 / 217                         & 41    & 96.1 / 222  \\
      EMPMP                                & 447K  & 93.1 / 216                   & 213   & 94.2 / 217  \\
      MotionConformer                      & 9.23M & \textbf{88.6} / \textbf{201} & 909   & \textbf{88.8} / \textbf{201} \\
      \bottomrule
   \end{tabular}
   \vspace{6pt}
   \caption{Comparison of single person forecasting on \textit{CMU-MoCap} dataset. The models were trained to predict a maximum time-range of $1000ms$, the forecast error was measured at the timesteps $400ms$ and $1000ms$. }
   \label{tab:abs_mocap1}
\end{table}

The second experiment in Table~\ref{tab:abs_mocap3} evaluates the models on a longer time-range of $3000\,ms$. Especially for systems with longer reaction times, like robots or cars, longer forecasts can help to plan their actions better.
Here a larger gap between the models can be observed.
Specifically at the longer time-range, \textit{DeformMLP} also seems to suffer from the too short $10$ possible input timesteps, whereas the other models could predict the motion from $180$ input frames. 
Apart from that, the relative performance between the models is similar to the previous experiments again.

\begin{table}[H]
   \centering
      \fontsize{7pt}{9pt}\selectfont
      \begin{tabular}{|l|c|c|r|c|}
         \toprule
         Method & Size & \makecell{MPJPE \\ 1000/2000/3000} & FPS & \makecell{FADE \\ 1000/2000/3000} \\
         \midrule
         \textit{Repeat last frame}           & -     & 356 / 562 / { } 706 & - { } & 356 / 562 / { } 706 \\
         \textit{Last delta average}          & -     & 373 / 773 / 1198    & - { } & 373 / 773 / 1198 \\
         \textit{Ridge Regression}            & -     & 276 / 507 / { } 692 & - { } & 276 / 507 / { } 692 \\
         DeformMLP                            & 1.29M & 241 / 454 / { } 656 & 226 & 242 / 455 / { } 657 \\
         JRTransformer                        & 4.38M & 255 / 446 / { } 602 & 562 & 255 / 446 / { } 602 \\
         EMPMP                                & 3.91M & 233 / 385 / { } 489 & 201 & 234 / 386 / { } 490 \\
         EqMotion                             & 3.42M & 218 / 354 / { } 469 & 13  & 235 / 368 / { } 481 \\
         MotionConformer                      & 9.23M & \textbf{211} / \textbf{351} / { } \textbf{462} & 777 & \textbf{211} / \textbf{351} / { } \textbf{462} \\
         \bottomrule
      \end{tabular}
   \vspace{6pt}
   \caption{Comparison of single person forecasting on \textit{CMU-MoCap} dataset. The models were trained to predict a maximum time-range of $3000ms$, the forecast error was measured at $1000ms$, $2000ms$ and $3000ms$. Again, twice the output duration was used as input duration. The reason for the difference of \textit{repeat last frame} method at timestep~$1000ms$ to the table before is the windowing of the dataset. Here, the window is much larger, resulting in fewer samples, because the windows start later and end earlier.}
   \label{tab:abs_mocap3}
\end{table}

\subsection{Multi-person Forecasts}

The previous benchmarks in this work focused on single-person forecasting.
In many applications, like human-robot interactions, there is only one main actor, or the persons are unrelated to others.
In such cases, the forecasts can be run independently per person.
However, in some cases the persons can interact closely and influence each other's movements.
Therefore, the following small experiment is intended to show the support for multi-person inputs.

In the previous works of \textit{JRTransformer}~\cite{xu2023joint}, and \textit{EMPMP}~\cite{zheng2025efficient} the multi-person dataset \textit{CMU-MoCap} followed the preprocessing of \textit{MRT}~\cite{wang2021multi}, including the scale problems, and partially missing preprocessing source-codes explained in more detail in Appendix~\ref{app:audit_details}.
Besides that, the multi-person interactions are partially synthetic, because the dataset combines different sequences to three person sequences, as mentioned earlier as well.
Therefore, \textit{CHi3D}~\cite{fieraru2020three} is used instead for the experiment. It shows two persons interacting closely with each other, and has a recording setup very similar to the one of the \textit{Human3.6M} dataset, using four cameras in the room edges.

The results in the upper part of Table~\ref{tab:chi3d} show that the single-person models of \textit{EqMotion} and \textit{MotionConformer} can successfully be extended to multi-person forecasts.
Technically the model inputs just need to be extended to $26$ instead of the original $13$ joints.
The general idea behind this simple approach is, that if the model can learn some connection between arm and leg movements of one person from input coordinates, it should be able to learn some connection between two hands of different persons as well.
In the paper of \textit{IAFormer}~\cite{xiao2024multi} this dataset was already used as well, and a \textit{MPJPE} of $218\,mm$, and $233\,mm$ for \textit{TBiFormer} were reported.
The results could not be replicated though, for unknown reasons there was no notable learning progress.
However, the reported \textit{MotionConformer} error is lower than their original published results.

\begin{table}[H]
   \centering
   \fontsize{8pt}{9pt}\selectfont
   \begin{tabular}{@{}|l|c|c|r|r|c|@{}}
      \toprule
      Method                                 & Size  & MPJPE & FPS  & FCE  & FADE \\
      \midrule
      \textit{Repeat last frame}             & -     & 272 & - { } & - { } &  272 \\
      \textit{Last delta average}            & -     & 318 & - { } & - { } &  318 \\
      \textit{Ridge Regression}              & -     & 277 & - { } & - { } &  277 \\
      EqMotion                               & 3.37M & 234   & 16   & 125  &  249 \\
      EMPMP                                  & 314K  & 227   & 209  &  10  &  228 \\
      MotionConformer                        & 10.2M & 207   & 853  &  2   &  207 \\
      \midrule
      MC \textit{(chi3d+h36m)}               & 10.2M & 191   & 853  &  2   &  191 \\
      \bottomrule
   \end{tabular}
   \vspace{6pt}
   \caption{Forecasting close two-person interactions on \textit{CHi3D}, errors at timestep $1000\,ms$.}
   \label{tab:chi3d}
   \vspace{-3pt}
\end{table}

\looseness=-1
Since the dataset is very small (the trainset only has around 0:13h of movements), especially the larger models suffer a bit from the data sparsity.
Besides the iterative finetuning on live data, which will be explored in more detail in the next section, one also could enhance the training dataset with synthetically created pair motions, or by adding similar single person motions to it.
In a short experiment, which simply added the \textit{Human3.6M} dataset to the training sequences, the results of the \textit{MotionConformer} in the lower part of Table~\ref{tab:chi3d} notably improved. 
This also shows transfer capabilities of the model from single to multi-person motions.

\subsection{Further Suggestions}

As a last note in this section, as a suggestion for future work, it should be beneficial to also add positions of objects and obstacles into the model inputs.
The current forecasting datasets are normally object and obstacle free, but in realistic environments this is seldom the case. 
Often humans interact with objects as well, like in a soccer game in which the ball will notably guide the future movement direction, or try to avoid obstacles, like in buildings with doors.
But this is out of scope for this work, and will be left for future research.


\section{Noisy Pose Evaluation}
\label{chap:noisy}

While recent pose forecasting methods report strong results on clean motion-capture data, deployment systems often rely on poses estimated from images.
The resulting detector noise can introduce spatially structured and temporally correlated errors that are not captured by clean-label evaluation.

\subsection{Previous Approaches}

With \cite{cui2021towards} and \cite{saadatnejad2023generic} there already are two works that investigated the impact of artificially generated noise.
Both of them randomly dropped some of the joints and also added Gaussian noise to the joint positions.
Only \textit{DePOSit}~\cite{saadatnejad2023generic} open-sourced their code for further usage.
Dropping random joints is a type of noise that normally does not occur in markerless settings, because most pose estimators always attempt to estimate joints that are not visible, without explicitly marking those estimated joints (which would allow dropping them afterward).
Regarding the Gaussian noise, across multiple frames (as in the model inputs) random noise averages around the true location, while pose estimators rather have more structural errors (like forearm too short in all frames) in which the noise doesn't follow a Gaussian distribution.
Due to those reasons, this work will investigate a more realistic type of error, namely the one caused by a preceding pose estimation network.

\vspace{-4pt}
\subsection{Detector-Noisy Data}
\vspace{-2pt}

To simplify the comparison between the performance with and without such noisy joint labels, a dataset is needed which also contains ground-truth labels.
While the authors of~\cite{sampieri2022pose} already have created a pose forecasting dataset with predicted joint labels (using \textit{VoxelPose}~\cite{tu2020voxelpose}), it has no ground-truth labels.
Furthermore, it contains almost no global movements of the persons as they are standing at a fixed position in front of a robot and mostly only move their arms.

Therefore, a new variant of the \textit{Human3.6M} dataset is generated.
The joints are predicted by a state-of-the-art multi-view multi-person model, namely \textit{RapidPoseTriangulation}~\cite{rapidtriang}.
It achieves strong zero-shot performance on the previously unseen \textit{Human3.6M} dataset of $52mm$ \textit{MPJPE}.
In the very rare cases ($0.1\%$) that the person was not detected, or only with a score below a threshold (of~$0.1$ in a range from~$0$-$1$), the joints of this frame were replaced with the joints of a valid previous frame, mirroring a simple real-time tracking policy.
In case multiple persons were detected, the one with the best score was kept.
The pose prediction was very fast, with an average of $160$~\textit{FPS} on a single Nvidia-RTX4080 GPU.
For consistency with the main benchmark, the data is subsampled to $25$ frames per $1000ms$ again.

\looseness=-1
The reason for choosing a multi-view pose estimator is that for most practical applications, multi-view systems are expected to be a notably better choice than single-view systems.
While they need some extra setup effort, this is comparably small, and avoiding/reducing errors in the input skeleton sequences, caused for example by occlusions, depth ambiguities, or image-to-pose estimation errors in one view, far outweighs the small additional calibration effort. 
Especially in professional or semi-professional setups in which a high pose and forecast accuracy is important.
This was investigated in more depth in \cite{skarimva}, which showed, for the related task of skeleton-based action recognition, that by switching from single-view to multi-view poses, the action classification error of multiple state-of-the-art methods could be significantly reduced, and also described why they are preferable to lifted 2D-to-3D skeletons from single-view inputs.
Since the tasks are very similar, it seems reasonable to argue that those results transfer to skeleton-based motion forecasting as well.

\subsection{Noise Adaptation Experiments}

To approximate a deployment scenario in which a forecasting model is available before being adapted to a new application setup, the models are first pretrained on a mixture of external motion-capture datasets (here \textit{CMU-MoCap}~\cite{mocap} and \textit{BMLmovi}~\cite{BMLmovi}, \textit{BMLrub}~\cite{BMLrub}, \textit{KIT}~\cite{KIT_Dataset} from the \textit{AMASS}~\cite{AMASS:2019} collection, which together contain around \mbox{$11$:$30h$} of motions).
The 13 joints from before are used again, but in the case of \textit{CMU-MoCap} the \textit{nose} is replaced with the \textit{upper-head} joint position since no \textit{nose} joint exists in this dataset.
The original frame-rates of those datasets are $120fps$ or $30fps$, but to match the target dataset, forecast is done with $25$ steps.
This changes the effective motion speed for some source datasets, but it is expected that the model should still be able to learn the general motion patterns.

Before evaluating detector noise, Table~\ref{tab:mh_10} reports clean-input zero-shot transfer to \textit{Human3.6M}.
Generalization to unseen datasets is a central requirement for practical forecasting models, and the results are close to the supervised \textit{Human3.6M} benchmark in Table~\ref{tab:abs_h36m}.
This indicates that the pretrained models have strong cross-dataset generalization capability.
Furthermore, the results provide a clean-input reference for measuring the degradation caused by detector-generated poses.

\begin{table}[H]
   \centering
   \fontsize{8pt}{9pt}\selectfont
   \begin{tabular}{|l|ccc|}
      \toprule
      Method & \multicolumn{3}{c|}{MPJPE} \\
               & $40ms$ & $400ms$ & $1000ms$ \\
      \midrule
      EMPMP             & 4.3 &  59.7 & 165 \\
      EqMotion          & 4.0 &  52.0 & 156 \\
      MotionConformer   & 4.0 &  51.6 & 149 \\
      \bottomrule
   \end{tabular}
   \vspace{6pt}
   \caption{Zero-shot transfer capabilities from a large mixture of datasets to \textit{Human3.6M}, at the given timesteps.}
   \label{tab:mh_10}
\end{table}

\begin{table}[!b]
   \centering
   \fontsize{8pt}{9pt}\selectfont
   \begin{tabular}{|l|cc|cc|}
      \toprule
      Method & \multicolumn{2}{c|}{detector target} & \multicolumn{2}{c|}{ground-truth target} \\
             & { } $400ms$ & $1000ms$ & { }{ }$400ms$ & $1000ms$ \\
      \midrule
      EqMotion                                 & { }  513 &  842       & { }{ }{ }532 & 856 \\
      EMPMP                                    & { }  100 &  215       & { }{ }{ }123 & 229 \\
      MotionConformer                          & { } 94.2 &  214       & { }{ }{ }118 & 228 \\
      \bottomrule
   \end{tabular}
   \vspace{6pt}
   \caption{Influence of skeleton joints generated by a 3D-pose-estimation network on model performance, measured as MPJPE in millimeters at the given timesteps. The left section uses the network predictions as inputs and targets (which would be the measurable error in live systems), and the right one the predictions for input and ground-truth labels as targets (which would be the real error).}
   \label{tab:zs_10}
\end{table}

Table~\ref{tab:zs_10} evaluates the same pretrained models when the input poses are produced by the 3D pose-estimation network.
The detector-target columns compare forecasts against future detector-generated poses, corresponding to the error measurable in a deployed system without motion-capture ground truth. The ground-truth-target columns compare the same forecasts against clean future poses, measuring the true error with respect to the actual human motion, including the upstream pose-estimator error (here on average $63\,mm$ at time zero).
Compared with the clean-input reference in Table~\ref{tab:mh_10}, detector-generated inputs cause a substantial performance degradation, and the true error is consistently higher than the detector-target error observable from detector outputs alone.

\looseness=-1
Since detector outputs can be collected without additional motion-capture labels, the forecasting model can be adapted to the target application setup using detector-generated pose sequences as pseudo-labels.
The upper block of Table~\ref{tab:ft_10} reports this adaptation for the pretrained \textit{MotionConformer}.
This setting simulates a deployed system that has collected observations from its target environment~(here for around~$1$:$55h$) and finetunes the forecasting model with them.
Following this approach, the detector-target error improves strongly~(by $25.7\%$), and the ground-truth-target error also decreases~(by $22.4\%$), indicating that pseudo-label adaptation reduces part of the detector-induced degradation.
Although training and evaluating on clean motion-capture data still gives the best clean-input result~($143\,mm$ at $1000\,ms$), the adapted detector-trained model remains competitive under detector-target evaluation~($159\,mm$ at $1000\,ms$), suggesting that detector-generated pseudo-labels can be a viable training signal when clean labels are unavailable.
Additional experiments with artificial Gaussian-noise pretraining are reported in Appendix~\ref{app:noise_pretraining}, showing that pretraining with artificial noise can improve robustness to detector-generated inputs but does not replace adaptation on the target detector outputs.

\begin{table}[H]
   \centering
   \fontsize{8pt}{9pt}\selectfont
   \begin{tabular}{|l|cc|cc|}
      \toprule
      Method & \multicolumn{2}{c|}{detector target} & \multicolumn{2}{c|}{ground-truth target} \\
             & { } $400ms$ & $1000ms$ & { }{ }$400ms$ & $1000ms$ \\
      \midrule
      MotionConformer                             & { } 66.9 & 159 & { }{ }{ } 94.8 & 177    \\
      \midrule
      EMPMP                                       & { } 76.4 & 179 & { }{ }{ }104 &  195 \\
      EqMotion                                    & { } 68.5 & 168 & { }{ }{ }96.1 &  185    \\
      MotionConformer                             & { } 69.1 & 165 & { }{ }{ }98.3 &  183    \\
      \bottomrule
   \end{tabular}
   \vspace{6pt}
   \caption{Improvements through pseudo-label training. Finetuning the pretrained model on noisy predictions in the upper part, training all models from scratch in the lower part. Columns are separated the same as in Table~\ref{tab:zs_10}}
   \label{tab:ft_10}
\end{table}

Because finetuning implementations were not available for the other two methods, the lower block of Table~\ref{tab:ft_10} reports the performance after training all models from scratch on the detector-generated dataset.
In terms of \textit{MPJPE} the results of \textit{EqMotion} and \textit{MotionConformer} are similar, but \textit{MotionConformer} retains the runtime advantage observed in the previous section.
The comparison between the two \textit{MotionConformer} rows shows that pretraining on a large motion dataset is beneficial, although the final accuracy improvement over from-scratch training is moderate.
Finetuning also converges faster, and a generic pretrained model is typically required before target-environment observations in a practical deployment scenario can be collected.


\section{Limitations}

The unified \textit{Human3.6M} benchmark is intended as a controlled reference protocol for future absolute pose forecasting evaluations.
Given the reproducibility and comparability issues found in existing benchmarks, it is proposed as a more reliable basis for future comparisons rather than merely as an additional leaderboard.
At the same time, its scope is protocol-specific: the primary benchmark uses a 13-joint, single-person, absolute-coordinate representation together with fixed preprocessing, windowing, and metric choices.
Results should therefore not be interpreted as numerically interchangeable with protocols that use different joint sets, coordinate systems, or evaluation metrics.
Additional experiments on \textit{CMU-MoCap} and \textit{CHI3D}, as reported in Section~\ref{app:more_benchs}, can provide complementary cross-dataset evaluations and should be used alongside the primary \textit{Human3.6M} protocol to assess whether observed performance trends generalize beyond a single dataset.

The benchmarks generally follow the common practice of the evaluated literature and report one training run per method.
This makes a broad audit of many third-party implementations feasible, but it does not measure training variability of all methods.
Such scores are thus best interpreted as reproducible artifact-level comparisons under the unified protocol.
For the best three methods, ablation experiments with multiple training runs with different random seeds were therefore performed and reported in Appendix~\ref{app:ablations}, and show a low variance of the reported scores.
Even across the other datasets in Section~\ref{app:more_benchs} (which at the same time use different seeds) the ranking stays the same, which provides further support for the overall ranking stability, although it does not establish it conclusively.

The cross-domain baseline study is also not an exhaustive architecture search, but rather tests whether mature generic sequence models are missing from current baseline comparisons.
For the proposed \textit{MotionConformer} models, trained checkpoints are released together with the evaluation code, making the reported artifact performance directly reproducible.
For third-party methods, reproducibility depends on the artifacts released by the original authors and on the compatibility of their training code with the unified protocol.
Future benchmark reports should therefore additionally include seed variation results when the claim concerns method-level superiority rather than the performance of a released checkpoint, especially when numerical margins are comparatively small.

The detector-noisy \textit{Human3.6M} variant measures robustness to one multi-view pose-estimation pipeline.
The selected detector was reported in its original paper to perform best across multiple datasets that were unseen during its training, making it a strong representative source of detector-induced noise.
Nevertheless, other pose estimators may induce different spatial and temporal error distributions.
Additionally, most commonly used datasets are captured in controlled laboratory settings, that might not perfectly reflect the conditions of real-world deployment. 
Therefore future work should evaluate whether the observed degradation and recovery behavior generalizes across detector families, and whether it persists under less controlled capture conditions.
However, such experiments are outside the scope of this work.
The proposed protocols nevertheless establish a reproducible and comparable framework for systematically evaluating the effects of detector noise under controlled conditions, while providing a common basis for future extensions.


\section{Conclusion}

This work re-examined human pose forecasting with a focus on evaluation reliability, model assumptions, and deployment realism.
The reproducibility audit uncovered several issues in prior evaluation code, preprocessing pipelines, artifact availability, and metric definitions.
By re-evaluating recent approaches under a unified absolute-pose protocol, this work provides a more controlled reference point for future comparisons.

The results also question whether progress in pose forecasting should be attributed primarily to specialized forecasting architectures.
By viewing pose forecasting as a sequence-to-sequence problem, architectures originally developed for speech recognition were adapted to this setting.
The adapted \textit{Conformer} models, and especially the optimized \textit{MotionConformer}, obtained the lowest reported errors under the unified protocol while maintaining high inference throughput.
This suggests that mature generic sequence models should be treated as serious baselines, and that architectural claims in pose forecasting should be evaluated under stricter and more comparable protocols.

To move evaluation closer to deployment conditions, this work introduced latency-aware metrics and a paired clean/detector-noisy \textit{Human3.6M} variant.
The noisy-input experiments show that models trained and evaluated only on clean motion-capture data can suffer a substantial performance drop when their inputs come from a pose estimator.
At the same time, adapting the forecasting model on detector-generated pose sequences can recover part of this degradation without requiring additional ground-truth annotations.

Overall, this work shows that rethinking human pose forecasting requires more than proposing new architectures.
Reliable progress depends on reproducible artifacts, clearly specified preprocessing, consistent metrics, strong generic baselines, and evaluation under realistic input conditions.
To support future work in this direction, all code, preprocessing scripts, trained \textit{MotionConformer} checkpoints, and evaluation tools are made publicly available.


\clearpage
\newpage

{
\small
\bibliographystyle{ieeenat_fullname}
\bibliography{egbib}

@String(ICCV= {Int. Conf. Comput. Vis.})

@String(ECCV= {Eur. Conf. Comput. Vis.})

@String(ICPR = {Int. Conf. Pattern Recog.})

@String(ICASSP=	{ICASSP})

@String(AAAI = {AAAI})

@String(ICCV  = {ICCV})

@String(ECCV  = {ECCV})

@String(ICPR  = {ICPR})

@article{
  rapidtriang,
  title={{RapidPoseTriangulation: Multi-view Multi-person Whole-body Human Pose Triangulation in a Millisecond}},
  author={Bermuth, Daniel and Poeppel, Alexander and Reif, Wolfgang},
  journal={arXiv preprint arXiv:2503.21692},
  year={2025}
}

@inproceedings{skarimva,
  title={Skarimva: Skeleton-based Action Recognition is a Multi-view Application},
  author={Bermuth, Daniel and Poeppel, Alexander and Reif, Wolfgang},
  booktitle={2026 IEEE 20th International Conference on Automatic Face and Gesture Recognition (FG)},
  pages={1--5},
  year={2026},
  organization={IEEE}
}

@inproceedings{martinez2017human,
  title={{On human motion prediction using recurrent neural networks}},
  author={Martinez, Julieta and Black, Michael J and Romero, Javier},
  booktitle={{Proceedings of the IEEE conference on computer vision and pattern recognition}},
  pages={2891--2900},
  year={2017}
}

@inproceedings{martinez2017simple,
  title={{A simple yet effective baseline for 3d human pose estimation}},
  author={Martinez, Julieta and Hossain, Rayat and Romero, Javier and Little, James J},
  booktitle={{Proceedings of the IEEE international conference on computer vision}},
  pages={2640--2649},
  year={2017}
}

@inproceedings{wei2020his,
  title={{History Repeats Itself: Human Motion Prediction via Motion Attention}},
  author={Wei, Mao and Miaomiao, Liu and Mathieu, Salzemann},
  booktitle={{ECCV}},
  year={2020}
}

@inproceedings{ijcai2022p111,
  title     = {{MotionMixer: MLP-based 3D Human Body Pose Forecasting}},
  author    = {Bouazizi, Arij and Holzbock, Adrian and Kressel, Ulrich and Dietmayer, Klaus and Belagiannis, Vasileios},
  booktitle = {Proceedings of the Thirty-First International Joint Conference on
               Artificial Intelligence, {IJCAI-22}},
  publisher = {International Joint Conferences on Artificial Intelligence Organization},
  pages     = {791--798},
  year      = {2022},
  month     = {7},
}

@misc{sofianos2021spacetimeseparable,
     title={{Space-Time-Separable Graph Convolutional Network for Pose Forecasting}}, 
     author={Theodoros Sofianos and Alessio Sampieri and Luca Franco and Fabio Galasso},
     year={2021},
     eprint={2110.04573},
     archivePrefix={arXiv},
     primaryClass={cs.CV}
}

@article{guo2022back,
  title={{Back to MLP: A Simple Baseline for Human Motion Prediction}},
  author={Guo, Wen and Du, Yuming and Shen, Xi and Lepetit, Vincent and Xavier, Alameda-Pineda and Francesc, Moreno-Noguer},
  journal={arXiv preprint arXiv:2207.01567},
  year={2022}
}

@inproceedings{adeli2021tripod,
  title={{Tripod: Human trajectory and pose dynamics forecasting in the wild}},
  author={Adeli, Vida and Ehsanpour, Mahsa and Reid, Ian and Niebles, Juan Carlos and Savarese, Silvio and Adeli, Ehsan and Rezatofighi, Hamid},
  booktitle={{Proceedings of the IEEE/CVF International Conference on Computer Vision}},
  pages={13390--13400},
  year={2021}
}

@article{vendrow2022somoformer,
  title={{SoMoFormer: Multi-Person Pose Forecasting with Transformers}},
  author={Vendrow, Edward and Kumar, Satyajit and Adeli, Ehsan and Rezatofighi, Hamid},
  journal={arXiv preprint arXiv:2208.14023},
  year={2022}
}

@article{h36m_pami,
  author = {Ionescu, Catalin and Papava, Dragos and Olaru, Vlad and Sminchisescu,  Cristian},
  title = {{Human3.6M: Large Scale Datasets and Predictive Methods for 3D Human Sensing in Natural Environments}},
  journal = {IEEE Transactions on Pattern Analysis and Machine Intelligence},
  publisher = {IEEE Computer Society},
  volume = {36},
  number = {7},
  pages = {1325-1339},
  month = {jul},
  year = {2014}
}

@inproceedings{von2018recovering,
  title={{Recovering accurate 3d human pose in the wild using imus and a moving camera}},
  author={Von Marcard, Timo and Henschel, Roberto and Black, Michael J and Rosenhahn, Bodo and Pons-Moll, Gerard},
  booktitle={{Proceedings of the European conference on computer vision (ECCV)}},
  pages={601--617},
  year={2018}
}

@article{wang2021multi,
  title={{Multi-person 3D motion prediction with multi-range transformers}},
  author={Wang, Jiashun and Xu, Huazhe and Narasimhan, Medhini and Wang, Xiaolong},
  journal={Advances in Neural Information Processing Systems},
  volume={34},
  pages={6036--6049},
  year={2021}
}

@misc{mocap,
     title={{CMU Graphics Lab Motion Capture Database}},
     author={Carnegie Mellon University: http://mocap.cs.cmu.edu/}
}

@inproceedings{giuliari2021transformer,
  title={{Transformer networks for trajectory forecasting}},
  author={Giuliari, Francesco and Hasan, Irtiza and Cristani, Marco and Galasso, Fabio},
  booktitle={{2020 25th international conference on pattern recognition (ICPR)}},
  pages={10335--10342},
  year={2021},
  organization={IEEE}
}

@article{kothari2021human,
  title={{Human trajectory forecasting in crowds: A deep learning perspective}},
  author={Kothari, Parth and Kreiss, Sven and Alahi, Alexandre},
  journal={IEEE Transactions on Intelligent Transportation Systems},
  volume={23},
  number={7},
  pages={7386--7400},
  year={2021},
  publisher={IEEE}
}

@inproceedings{AMASS:2019,
  title={{AMASS: Archive of Motion Capture as Surface Shapes}},
  author={Mahmood, Naureen and Ghorbani, Nima and F. Troje, Nikolaus and Pons-Moll, Gerard and Black, Michael J.},
  booktitle = {The IEEE International Conference on Computer Vision (ICCV)},
  year={2019},
  month = {Oct},
  url = {https://amass.is.tue.mpg.de},
  month_numeric = {10}
}

@inproceedings{tu2020voxelpose,
  title={{Voxelpose: Towards multi-camera 3d human pose estimation in wild environment}},
  author={Tu, Hanyue and Wang, Chunyu and Zeng, Wenjun},
  booktitle={{Computer Vision--ECCV 2020: 16th European Conference, Glasgow, UK, August 23--28, 2020, Proceedings, Part I 16}},
  pages={197--212},
  year={2020},
  organization={Springer}
}

@inproceedings{ma2022progressively,
  title={{Progressively generating better initial guesses towards next stages for high-quality human motion prediction}},
  author={Ma, Tiezheng and Nie, Yongwei and Long, Chengjiang and Zhang, Qing and Li, Guiqing},
  booktitle={{Proceedings of the IEEE/CVF Conference on Computer Vision and Pattern Recognition}},
  pages={6437--6446},
  year={2022}
}

@inproceedings{li2022skeleton,
  title={{Skeleton-parted graph scattering networks for 3d human motion prediction}},
  author={Li, Maosen and Chen, Siheng and Zhang, Zijing and Xie, Lingxi and Tian, Qi and Zhang, Ya},
  booktitle={{European Conference on Computer Vision}},
  pages={18--36},
  year={2022},
  organization={Springer}
}

@inproceedings{saadatnejad2023generic,
  title={{A generic diffusion-based approach for 3D human pose prediction in the wild}},
  author={Saadatnejad, Saeed and Rasekh, Ali and Mofayezi, Mohammadreza and Medghalchi, Yasamin and Rajabzadeh, Sara and Mordan, Taylor and Alahi, Alexandre},
  booktitle={{2023 IEEE International Conference on Robotics and Automation (ICRA)}},
  pages={8246--8253},
  year={2023},
  organization={IEEE}
}

@inproceedings{cui2021towards,
  title={{Towards accurate 3d human motion prediction from incomplete observations}},
  author={Cui, Qiongjie and Sun, Huaijiang},
  booktitle={{Proceedings of the IEEE/CVF Conference on Computer Vision and Pattern Recognition}},
  pages={4801--4810},
  year={2021}
}

@inproceedings{xu2023eqmotion,
  title={{EqMotion: Equivariant Multi-agent Motion Prediction with Invariant Interaction Reasoning}},
  author={Xu, Chenxin and Tan, Robby T and Tan, Yuhong and Chen, Siheng and Wang, Yu Guang and Wang, Xinchao and Wang, Yanfeng},
  booktitle={{Proceedings of the IEEE/CVF Conference on Computer Vision and Pattern Recognition}},
  pages={1410--1420},
  year={2023}
}

@article{BMLrub,
	title        = {{Decomposing Biological Motion: {A} Framework for Analysis and Synthesis of Human Gait Patterns}},
	author       = {Troje, Nikolaus F.},
	year         = 2002,
	month        = sep,
	journal      = {Journal of Vision},
	volume       = 2,
	number       = 5,
	pages        = {2--2},
	doi          = {10.1167/2.5.2},
	month_numeric = 9
}

@inproceedings{KIT_Dataset,
	title        = {{The {KIT} whole-body human motion database}},
	author       = {C. {Mandery} and Ö. {Terlemez} and M. {Do} and N. {Vahrenkamp} and T. {Asfour}},
	year         = 2015,
	month        = jul,
	booktitle    = {2015 International Conference on Advanced Robotics (ICAR)},
	volume       = {},
	number       = {},
	pages        = {329--336},
	doi          = {10.1109/ICAR.2015.7251476},
	month_numeric = 7,
	event_place  = {Istanbul, Turkey}
}

@article{BMLmovi,
  title={{MoVi: A large multi-purpose human motion and video dataset}},
  author={Ghorbani, Saeed and Mahdaviani, Kimia and Thaler, Anne and Kording, Konrad and Cook, Douglas James and Blohm, Gunnar and Troje, Nikolaus F},
  journal={Plos one},
  volume={16},
  number={6},
  pages={e0253157},
  year={2021},
  publisher={Public Library of Science San Francisco, CA USA}
}

@inproceedings{sampieri2022pose,
  title={{Pose forecasting in industrial human-robot collaboration}},
  author={Sampieri, Alessio and di Melendugno, Guido Maria D’Amely and Avogaro, Andrea and Cunico, Federico and Setti, Francesco and Skenderi, Geri and Cristani, Marco and Galasso, Fabio},
  booktitle={{European Conference on Computer Vision}},
  pages={51--69},
  year={2022},
  organization={Springer}
}

@article{shi2024gradient,
  title={{Gradient multi-foci networks for 3D skeleton-based human motion prediction}},
  author={Shi, Junyu and Zhong, Jianqi and He, Zhiquan and Cao, Wenming},
  journal={Neural Computing and Applications},
  volume={36},
  number={24},
  pages={14627--14642},
  year={2024},
  publisher={Springer}
}

@inproceedings{xu2023auxiliary,
  title={{Auxiliary tasks benefit 3d skeleton-based human motion prediction}},
  author={Xu, Chenxin and Tan, Robby T and Tan, Yuhong and Chen, Siheng and Wang, Xinchao and Wang, Yanfeng},
  booktitle={{Proceedings of the IEEE/CVF international conference on computer vision}},
  pages={9509--9520},
  year={2023}
}

@inproceedings{xiao2024multi,
  title={{Multi-person Pose Forecasting with Individual Interaction Perceptron and Prior Learning}},
  author={Xiao, Peng and Xie, Yi and Xu, Xuemiao and Chen, Weihong and Zhang, Huaidong},
  booktitle={{European Conference on Computer Vision}},
  pages={402--419},
  year={2024},
  organization={Springer}
}

@inproceedings{wang2024gcnext,
  title={{Gcnext: Towards the unity of graph convolutions for human motion prediction}},
  author={Wang, Xinshun and Cui, Qiongjie and Chen, Chen and Liu, Mengyuan},
  booktitle={{Proceedings of the AAAI Conference on Artificial Intelligence}},
  volume={38},
  number={6},
  pages={5642--5650},
  year={2024}
}

@inproceedings{huang2024deformmlp,
  title={{DeformMLP: dynamic large-scale receptive field MLP networks for human motion prediction}},
  author={Huang, Haitao and Pun, Chi-Man and Li, Haolun and Liu, Mengqi and Xiong, Jian and Gao, Hao},
  booktitle={{ICASSP 2024-2024 IEEE International Conference on Acoustics, Speech and Signal Processing (ICASSP)}},
  pages={5200--5204},
  year={2024},
  organization={IEEE}
}

@inproceedings{xu2023joint,
  title={{Joint-relation transformer for multi-person motion prediction}},
  author={Xu, Qingyao and Mao, Weibo and Gong, Jingze and Xu, Chenxin and Chen, Siheng and Xie, Weidi and Zhang, Ya and Wang, Yanfeng},
  booktitle={{Proceedings of the IEEE/CVF International Conference on Computer Vision}},
  pages={9816--9826},
  year={2023}
}

@inproceedings{peng2023trajectory,
  title={{Trajectory-aware body interaction transformer for multi-person pose forecasting}},
  author={Peng, Xiaogang and Mao, Siyuan and Wu, Zizhao},
  booktitle={{Proceedings of the IEEE/CVF conference on computer vision and pattern recognition}},
  pages={17121--17130},
  year={2023}
}

@inproceedings{kriman2020quartznet,
  title={{Quartznet: Deep automatic speech recognition with 1d time-channel separable convolutions}},
  author={Kriman, Samuel and Beliaev, Stanislav and Ginsburg, Boris and Huang, Jocelyn and Kuchaiev, Oleksii and Lavrukhin, Vitaly and Leary, Ryan and Li, Jason and Zhang, Yang},
  booktitle={{ICASSP 2020-2020 IEEE International Conference on Acoustics, Speech and Signal Processing (ICASSP)}},
  pages={6124--6128},
  year={2020},
  organization={IEEE}
}

@article{gulati2020conformer,
  title={{Conformer: Convolution-augmented transformer for speech recognition}},
  author={Gulati, Anmol and Qin, James and Chiu, Chung-Cheng and Parmar, Niki and Zhang, Yu and Yu, Jiahui and Han, Wei and Wang, Shibo and Zhang, Zhengdong and Wu, Yonghui and others},
  journal={arXiv preprint arXiv:2005.08100},
  year={2020}
}

@article{kim2022squeezeformer,
  title={{Squeezeformer: An efficient transformer for automatic speech recognition}},
  author={Kim, Sehoon and Gholami, Amir and Shaw, Albert and Lee, Nicholas and Mangalam, Karttikeya and Malik, Jitendra and Mahoney, Michael W and Keutzer, Kurt},
  journal={Advances in Neural Information Processing Systems},
  volume={35},
  pages={9361--9373},
  year={2022}
}

@article{zheng2025efficient,
  title={{Efficient Multi-Person Motion Prediction by Lightweight Spatial and Temporal Interactions}},
  author={Zheng, Yuanhong and Yu, Ruixuan and Sun, Jian},
  journal={arXiv preprint arXiv:2507.09446},
  year={2025}
}

@article{
  scribosermo,
  title={{Scribosermo: Fast Speech-to-Text models for German and other Languages}},
  author={Bermuth, Daniel and Poeppel, Alexander and Reif, Wolfgang},
  journal={arXiv preprint arXiv:2110.07982},
  year={2021}
}

@misc{DEPSPE,
	author={Mozilla},
	title={{Project DeepSpeech}},
	year={2021},
	url={https://github.com/mozilla/DeepSpeech},
	note={[accessed 26-February-2021]}
}

@inproceedings{jeong2024multi,
  title={Multi-agent long-term 3d human pose forecasting via interaction-aware trajectory conditioning},
  author={Jeong, Jaewoo and Park, Daehee and Yoon, Kuk-Jin},
  booktitle={Proceedings of the IEEE/CVF Conference on Computer Vision and Pattern Recognition},
  pages={1617--1628},
  year={2024}
}

@inproceedings{li2025component,
  title={Component-wise Self-Correction Network for Human Motion Prediction},
  author={Li, Jinkai and Wang, Jinghua and Wang, Xin and Yan, Liang and Xu, Yong},
  booktitle={ICASSP 2025-2025 IEEE International Conference on Acoustics, Speech and Signal Processing (ICASSP)},
  pages={1--5},
  year={2025},
  organization={IEEE}
}

@inproceedings{fieraru2020three,
  title={{Three-dimensional reconstruction of human interactions}},
  author={Fieraru, Mihai and Zanfir, Mihai and Oneata, Elisabeta and Popa, Alin-Ionut and Olaru, Vlad and Sminchisescu, Cristian},
  booktitle={Proceedings of the IEEE/CVF Conference on Computer Vision and Pattern Recognition},
  pages={7214--7223},
  year={2020}
}
}

\clearpage
\appendix


\section{Related Work}
\label{app:relwork}

This section provides a more detailed overview of the aforementioned related work methods evaluated in the experiments of this paper. The selection covers the main approaches identified in the literature, with the majority having been published at top-tier conferences. The key architectural aspects of each method are described below.

\subsection{Relative Forecasting Approaches}

Most relative approaches build upon the source-codes from~\cite{martinez2017simple, martinez2017human} and reuse their preprocessed datasets and evaluation metrics.
The default dataset is \textit{Human3.6M}~\cite{h36m_pami}.
In the commonly used subset, 7~actors are performing 15~different actions.
The actors are represented as skeletons with 32~joints in total, which in the case of the preprocessed dataset, are represented as exponential maps, from which the 3D coordinates are computed.
From the seven actors, the subject~\textit{S11} is used for validation, and~\textit{S5} is used for testing.
The dataset was captured at a frame rate of $50Hz$, but usually every second frame is skipped, and only 25~future frames are predicted for the next second.

\begin{figure}[htpb]
   \centering
   \includegraphics[width=0.475\linewidth]{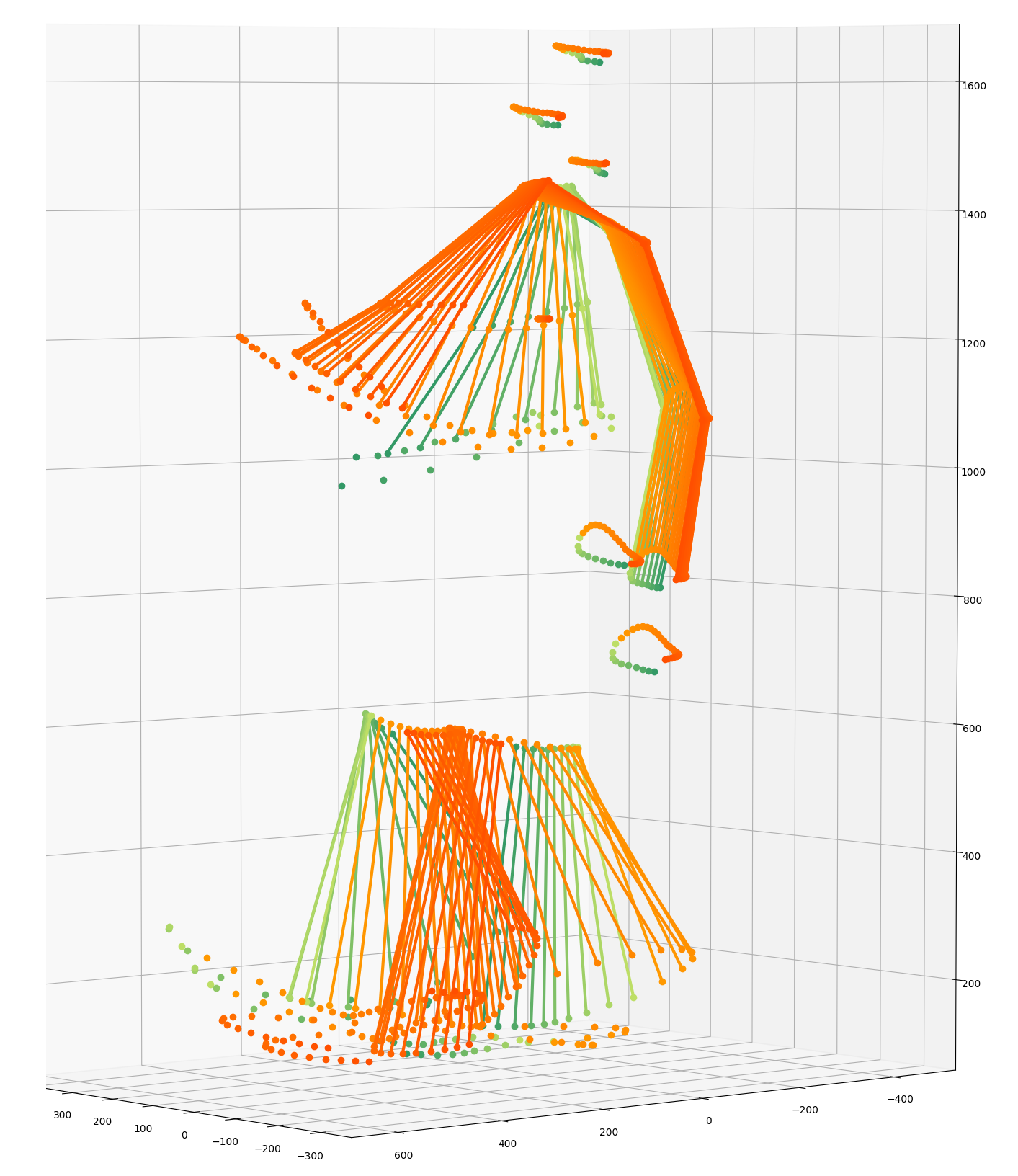}
   \caption{Example of a \textit{relative} pose forecast of a walking person. The green skeletons visualize the input sequence of the prediction model, and the red ones the predicted future poses. Only the predicted joints are visualized, therefore this person does not have a hip (it is fixed to the same spot).}
   \label{fig:rel_forecast}
\end{figure}

Out of the 32~joints 22~joints are selected for training (see Figure~\ref{fig:rel_forecast}), the others either contain the static hip joints or are duplicated joints.
To calculate the resulting \textit{mean per joint position error (MPJPE)} according to the default protocol, the 32~target joints are copied and the values of the 22~training joints are replaced with the model predictions.
Then the mean error over all 32~joints is calculated.
For each action 256~randomly sampled sequences are used.

The relative approaches investigated in this paper mostly use different network architectures. 
\textit{HisRepItself}~\cite{wei2020his} uses a motion-attention concept to attend to historical motion sub-sequences, which are then used by a graph convolution network to predict future poses. 
\textit{STSGCN}~\cite{sofianos2021spacetimeseparable} only uses a graph convolution network, which can separate between the space and time axis of the motions.
\textit{SPGSN}~\cite{li2022skeleton} splits the skeleton into separate parts and analyzes both, parts and the complete skeleton model to extract features and predict future poses.
\textit{PGBIG}~\cite{ma2022progressively} uses a multi-stage graph network in which initial guesses are progressively improved until the final output.
\textit{MotionMixer}~\cite{ijcai2022p111} relies solely on multi-layer perceptrons, which alternate between using spatial and temporal channels as inputs. 
\textit{siMLPe}~\cite{guo2022back} uses a similar, but much simpler architecture, which in its core has only alternating fully-connected and layer-norm layers.
\textit{EqMotion}~\cite{xu2023eqmotion} distinguishes between geometric and pattern features and additionally creates an interaction graph for the inputs. The model is based on fully-connected layers and has multiple connections between the geometric and pattern layers.
\mbox{\textit{DePOSit}}~\cite{saadatnejad2023generic} uses a diffusion model to simultaneously predict future poses and de-noise them. 
In some of their experiments, the prediction was done in two steps with two models, one for the short-term future as an intermediate step, and the second that uses the predictions from the first as input for the complete prediction.
Its architecture was directly designed to handle noisy inputs as well.
\textit{AuxFormer}~\cite{xu2023auxiliary} uses a transformer-based architecture, which includes auxiliary tasks in its training, to improve the forecasting performance.
Instead of just forecasting future poses, it also has network branches designed to repair artificially added noise and dropout errors.
\textit{GMFnet}~\cite{shi2024gradient} developed a two-stage model to improve the focus of attention layers for better feature extraction.  
\textit{GCNext}~\cite{wang2024gcnext} improves the design of graph convolution layers to allow them to calculate features across space, time and channels (joints) simultaneously.
\textit{DeformMLP}~\cite{huang2024deformmlp} uses a multi-layer perceptron architecture that splits inputs into non-overlapping blocks which are also mixed across space, time and channel dimensions.

\subsection{Absolute Forecasting Approaches}

Later on, there have been attempts to calculate absolute pose motions instead of relative poses.
For applications requiring the full human movement, this has the benefit that the model outputs can be used directly, and do not need to be combined with models that focus on forecasting only the general motion trajectory, like~\cite{giuliari2021transformer, kothari2021human}.
An earlier approach for this is \textit{TRiPOD}~\cite{adeli2021tripod}, which also created a new benchmark.
For 3D joint predictions they use the \textit{3D Poses in the wild}~(3DPW) dataset~\cite{von2018recovering}, which consists of 60~action sequences in a real world setting. 
Another dataset often used for absolute pose forecasting is \textit{CMU-MoCap}~\cite{mocap}.
The actors have been captured by twelve cameras in a $3m\,$$\times$$\,8m$ room with a frame-rate of $120Hz$. Usually this rate is reduced to $15$ frames per second by skipping over frames.
In most scenes there is only one actor, but some contain a second actor as well.

\textit{SoMoFormer}~\cite{vendrow2022somoformer} was one of the earlier leading algorithms on both datasets and is also available open-source.
The algorithm uses a transformer-based method, which uses a learned joint and person embedding to allow movement predictions for multiple persons at the same time.
\mbox{\textit{TBiFormer}}~\cite{peng2023trajectory} splits the persons into sequences of body parts and then learns self- and inter-person relations using those part features.
\textit{JRTransformer}~\cite{xu2023joint} distinguishes between joint and relation features and has specialized layers for both, which are regularly fused with each other.
\textit{IAFormer}~\cite{xiao2024multi} focuses on the interaction between persons and tries to learn person-independent interaction features to especially improve multi-person forecasts.
\textit{T2P}~\cite{jeong2024multi} follows a coarse-to-fine approach, by first forecasting global trajectories, followed by conditioned local pose forecasts.
\textit{EMPMP}~\cite{zheng2025efficient} proposes a lightweight dual branch architecture to learn local and global motions together with a cross-level interaction module to share information between the two branches.


\section{Detailed Reproduction Notes}
\label{app:audit_details}

Before comparing all models on the same dataset, the first question was: \textit{Can the results of the approaches described in their respective papers be reproduced?}
What should have been a straightforward reproduction exercise instead revealed substantial issues in many of the audited methods.
This appendix provides the detailed evidence behind the audit summary in Table~\ref{tab:audit_summary}.
For each method, the reported result, available artifacts, reproduction outcome, and identified discrepancy are described.

\subsection{Relative Forecasting Approaches}

For relative forecasting methods, several approaches could be reproduced exactly or within small numerical differences, while others showed larger discrepancies.

\textit{STSGCN}~\cite{sofianos2021spacetimeseparable} had an error in their MPJPE calculation, which the authors mentioned on their linked repository.
The corrected results could be reproduced here.
\textit{MotionMixer}~\cite{ijcai2022p111} calculated both metrics, but also had an error in the old metric (at counting items for the average calculation), which could be corrected because they open-sourced the code, but it resulted in significantly lower performance.
The reason why \textit{SPGSN}~\cite{li2022skeleton} and \textit{EqMotion}~\cite{xu2023eqmotion} had a lower performance is unclear.
Similar reproduction difficulties were reported in public repository issues, but no resolved explanation was available at the time of the audit.
With \textit{DePOSit}~\cite{saadatnejad2023generic} and \textit{GMFnet}~\cite{shi2024gradient} the reason is unclear as well.
The results of \textit{EqMotion}, \textit{AuxFormer}, \textit{GMFnet}\,($400\,ms$) and \textit{DePOSit} could be reproduced with their published checkpoints though.
The repository of \textit{IAFormer}~\cite{xiao2024multi} did not contain the code for the \textit{Human3.6M} dataset, but their results on \textit{CMU-MoCap} could be replicated.
\textit{CSCNet}~\cite{li2025component} has no published source code.
A comparison of all relative approaches, including the corrected results, can be found in Table~\ref{tab:rel_compare}. 
\textit{Res.\,{}Sup.}~\cite{martinez2017human} served as the basis on which the other works build upon.

\begin{table}[H]
   \centering
   \fontsize{8pt}{9pt}\selectfont
   \begin{tabular}{|l|c|c|}
      \toprule
      Method & MPJPE (400/1000ms) & Replicated \\
      \midrule
      \textit{Repeat last frame}                                               & 88.2 / 136.6          & - \\
      \textit{Ridge Regression}                                                & 76.5 / 128.2          & - \\
      Res.\,{}Sup. \cite{martinez2017human, wei2020his}     \hspace{07pt}'2017 & 88.3 / 136.6          & - \\
      IAFormer \cite{xiao2024multi}                         \hspace{18pt}'2024 & 50.8 / 130.4          & - \\
      MotionMixer \cite{ijcai2022p111} \dag                 \hspace{06pt}'2022 & 59.3 / 111.0          & 65.4 / 117.9 \\
      STSGCN \cite{sofianos2021spacetimeseparable} \dag     \hspace{15pt}'2021 & 38.3 / { } 75.6       & 67.5 / 117.0 \\
      GMFnet \cite{shi2024gradient}                         \hspace{22pt}'2024 & { } { } - { } / 102.3 & 64.1 / 114.1 \\
      SPGSN \cite{li2022skeleton}                           \hspace{24pt}'2022 & 58.3 / 109.6          & 62.0 / 112.3 \\
      HisRepItself \cite{wei2020his}                        \hspace{08pt}'2020 & 58.3 / 112.1          & 58.3 / 112.0 \\
      AuxFormer \cite{xu2023auxiliary}                      \hspace{11pt}'2023 & { } { } - { } / 107.0 & 61.7 / 111.6 \\
      PGBIG \cite{ma2022progressively}                      \hspace{24pt}'2022 & 58.5 / 110.3          & 58.8 / 110.4 \\
      EqMotion \cite{xu2023eqmotion}                        \hspace{16pt}'2023 & 55.0 / 106.9          & 57.8 / 109.8 \\
      GCNext \cite{wang2024gcnext}                          \hspace{22pt}'2024 & 56.4 / 108.7          & 57.5 / 109.8 \\
      CSCNet \cite{li2025component}                         \hspace{22pt}'2025 & 57.4 / 109.4          & -  \\
      siMLPe \cite{guo2022back}                             \hspace{23pt}'2022 & 57.3 / 109.4          & 57.6 / 109.6 \\
      DePOSit \cite{saadatnejad2023generic}                 \hspace{20pt}'2023 & { } { } - { } / 103.3 & 59.0 / 106.2 \\
      DeformMLP \cite{huang2024deformmlp}                   \hspace{07pt}'2024 & 57.9 / 105.5          & 57.5 / 105.3 \\
      \bottomrule
   \end{tabular}
   \vspace{6pt}
   \caption{Comparing recent approaches for \textit{relative} pose forecasting on \textit{Human3.6M} dataset, averaged over all actions. The \textit{mean per joint position error}~(MPJPE) is measured in millimeters at two frames 400\,ms and 1000\,ms in the future (using the same model for both timesteps). Approaches marked with a \dag{} had errors in their evaluation code which were fixed in the \textit{replicated} column.}
   \label{tab:rel_compare}
\end{table}

\subsection{Absolute Forecasting Approaches}

For absolute forecasting methods, the main reproducibility problems were metric definitions, scale conversions, and timestamp conventions.

In the absolute pose forecasting benchmark of \textit{TRiPOD}~\cite{adeli2021tripod} a new metric called \textit{Visibility-Ignored Metric}~(VIM) was developed, which in the case of no invisible joints, should behave like MPJPE for each timestep (\textit{" This metric is the simple MPJPE metric except that the invisible joints (if exist) are not penalized and are simply discarded by considering truth"}), but it has an error in its formula (see Equations~(\ref{eq:vim})~and~(\ref{eq:mpjpe})).
Instead of calculating the distance in 3D-space and averaging over the joints, the distance is calculated in 3*N-dimensional joint space.
It is measured in centimeters. To calculate VIM the train/test splits of 3DPW also need to be switched.

\begin{equation}
   \small
   VIM_{t} = \sqrt{\sum_{i=1}^{joints}\left(\sum_{k=1}^{3} (gt_{t,i,k} - pred_{t,i,k})^2\right)}
   \label{eq:vim}
\end{equation}



\looseness=-1
For comparison on the \textit{CMU-MoCap} dataset, it was stated in  \textit{SoMoFormer} that the same training recipe and evaluation metric was used as in \textit{MRT}~\cite{wang2021multi}.
In the \textit{MRT}-paper it was described that the error unit is $0.1$~meters, which already includes a dataset scaling factor of $0.6$, but following the source code of \textit{MRT}, a different unit has to be used.
The error has to be multiplied by $((10*3/1.8*si2m)/1.8)\approx0.5226$ to get a unit of $1$~meter, with \mbox{$si2m=((1.0 / 0.45) * 2.54 / 100.0)\approx0.0564$}, which following the dataset description, converts the scaled inches unit to meters.
Following the large difference between the results reported in the \textit{MRT}-paper, and the ones for the \textit{MRT}-model in the \textit{SoMoFormer}-paper, the latter likely used a different metric or scale, but this could not be validated, because the source code was not completely open-sourced. 
The same also applies to \textit{TBiFormer}~\cite{peng2023trajectory}, \textit{JRTransformer}~\cite{xu2023joint}, and \textit{EMPMP}~\cite{zheng2025efficient}.
Besides that, all papers reported different results for the same benchmark in their own replications, sometimes resulting in older models being better than the newer ones, or with large numeric error differences, which shows general replicability problems with those models (Table~\ref{tab:abs_compare_mocap}).
One reason could be that often only parts of the code are published, so some steps might have been reimplemented differently.
And for \textit{T2P}~\cite{jeong2024multi} the environment could not be set up completely, therefore it was not evaluated in the main sections.

\begin{table}[htpb]
   \begin{center}
      \fontsize{8pt}{9pt}\selectfont
      \begin{tabular}{|l|l|c|}
         \toprule
         Method & Reporting Paper & MPJPE (1000ms / 3000ms) \\
         \midrule
         HisRepItself & MRT \cite{wang2021multi}                       &  1.05 / 1.58  \\
                      & TBiFormer \cite{peng2023trajectory}            &  134 /  349  \\
                      & SoMoFormer \cite{vendrow2022somoformer}        &  0.50 /  1.42  \\
                      & JRTransformer \cite{xu2023joint}               &  14.9 /  30.7  \\
         \midrule
         MRT & MRT \cite{wang2021multi}                       &  0.79 / 1.22  \\
             & TBiFormer \cite{peng2023trajectory}            &  148 / 352  \\
             & SoMoFormer \cite{vendrow2022somoformer}        &  0.46 / 1.11  \\
             & JRTransformer \cite{xu2023joint}               & 9.6 / 21.8  \\
             & EMPMP \cite{zheng2025efficient}                &  89 /  189  \\

         \midrule
         TBiFormer & TBiFormer \cite{peng2023trajectory}            &  118 / 329  \\
                   & EMPMP \cite{zheng2025efficient}                &  98 / 218  \\
         \midrule
         SoMoFormer & SoMoFormer \cite{vendrow2022somoformer}        &  0.42 / 1.06 \\
                    & JRTransformer \cite{xu2023joint}              &  10.7 / 22.5  \\
                    & EMPMP \cite{zheng2025efficient}              &  84 / 202  \\
         \midrule
         JRTransformer & JRTransformer \cite{xu2023joint}            &  8.3 / 18.5  \\
                       & EMPMP \cite{zheng2025efficient}           &  88 / 206  \\
         \bottomrule
      \end{tabular}
   \end{center}
   \caption{Comparing reported results of recent approaches for \textit{absolute} pose forecasting on the \textit{CMU-MoCap} dataset proposed by \textit{MRT}~\cite{wang2021multi}. Even though the descriptions of the experiments are similar, the reported results differ substantially between papers, showing general replicability problems.}
   \label{tab:abs_compare_mocap}
\end{table}

The authors of \textit{TBiFormer} stated to use the preprocessing from \textit{MRT} as well, but if that is correct, at least one of their results uses wrong timestamps.
In the published source-code they used 50 input and 25 output frames for their 2s/1s experiment on \textit{CMU-MoCap}.
The base dataset has an original frame-rate of $120Hz$, which is reduced to $30Hz$ by \textit{MRT}.
If they did not interpolate the frames (the preprocessing code was not published), the predicted time range is too short.
The same problem could be found in the source-code of \textit{IAFormer} and \textit{T2P}, and the paper of \textit{GMFnet}.


\section{Converting relative to absolute Forecasters}
\label{app:convert}

To evaluate the released relative-pose forecasting implementations under the unified absolute-pose protocol, their input pipelines were replaced.
Instead of reusing the preprocessed data without global motion, which all approaches took from~\cite{martinez2017human}, the absolute poses were extracted directly from the \textit{Human3.6M} dataset.

The goal was to make the methods comparable at the level of data, coordinates, joint definitions, and metrics while avoiding full reimplementations of each architecture.
Whenever possible, models were trained with the same temporal setup: $50$ input frames and $25$ output frames for \textit{Human3.6M}.
Default hyperparameters from the released implementations were otherwise retained.

Some methods required compatibility modifications.
For \textit{DePOSit} the single-stage architecture without intermediate timesteps is used, similar to its original experiments with larger datasets, to significantly speed up training and inference (about $2\times$).
\textit{GMFnet} and \textit{DeformMLP} had the $22$ joints of the old dataset deeply integrated into their architecture, so to match their joint number, the $13$ main joints are padded to $22$.
For the \textit{MPJPE} calculation those padded joints were removed again from the prediction.
For \textit{DeformMLP}, the original 10-frame input length was retained because the 50-frame variant substantially increased model size and diverged in all attempted runs.

These modifications should be interpreted as minimal compatibility changes needed to evaluate available implementations under a shared protocol, not as an architecture-only comparison in which all methods were independently re-engineered to identical input assumptions.


\section{FADE Reasoning}
\label{app:fade}

The \textit{forecast after delay error (FADE)} is intended as a latency-aware approximation for fixed-horizon forecasting models.
If a model requires an inference time of~$\Delta t$, then a forecast requested for horizon~$t$ is only available after part of that horizon has already elapsed.
In a deployment system, the relevant prediction is therefore closer to horizon $t+\Delta t$ than to horizon $t$.
An exact evaluation of this delayed horizon would require each model to output predictions at delay-specific horizons, or would require retraining the models with longer output sequences.
\textit{FADE} avoids this additional retraining by estimating the error at the delayed horizon from the error measured at the original horizon.

This approximation assumes that the \textit{MPJPE} grows approximately linearly in a small neighborhood after the evaluated timestep.
Importantly, this is a local approximation around each forecast horizon, not a claim that the full \textit{MPJPE} curve is globally linear over the complete forecast range.
For a horizon $t$ and delay $\Delta t$, \textit{FADE} extrapolates the delayed error by extending the linear trend from zero error at time $0$ to the measured error $MPJPE_t$ at time $t$.

\begin{figure}[htpb]
   \includegraphics[width=\linewidth]{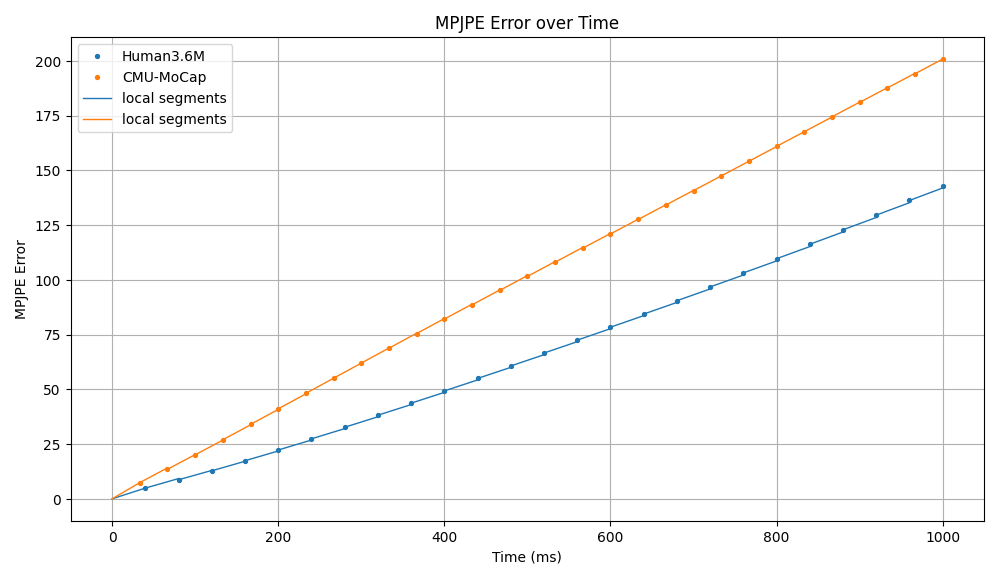}
   \vspace{-6pt}
   \caption{Measured \textit{MPJPE} over time for \textit{MotionConformer}, together with local one-step extrapolation segments used to validate the \textit{FADE} approximation. Each segment starts at a measured timestep and follows the linear increase assumed by \textit{FADE}. The small vertical offset between the segment endpoint and the next measured \textit{MPJPE} value indicates the local approximation error.}
   \label{fig:fade_mpjpe}
\end{figure}

Figure~\ref{fig:fade_mpjpe} visualizes this assumption for \textit{MotionConformer}.
The measured \textit{MPJPE} values are shown together with short extrapolation segments.
Each segment starts at a measured timestep and extends by one forecast step using the local linear increase that \textit{FADE} would assume at that point.
The vertical difference between the end of the segment and the next measured \textit{MPJPE} value therefore measures the local discrepancy between the \textit{FADE} approximation and the actual forecast error curve.

For \textit{Human3.6M} with $25$ steps per second, this local discrepancy has a mean absolute error of $0.70\,mm$, and a maximum absolute error of $1.07\,mm$.
For \textit{CMU-MoCap} with $30$ steps per second, the mean absolute error is $0.16\,mm$, and the maximum absolute error is $0.65\,mm$.
These errors are small compared with the \textit{MPJPE} values reported in the main experiments, supporting the use of \textit{FADE} as a latency-aware approximation for the evaluated settings.

\FloatBarrier
\section{Scriboora Details}
\label{app:scriboora}

This appendix provides implementation details for the cross-domain sequence baselines used in Section~\ref{sec:cdsb}.
The purpose of these models is not to claim that speech recognition is uniquely suited to pose forecasting, but to test whether mature sequence-to-sequence architectures from a neighboring domain already form strong baselines under the unified protocol.
Speech-to-text models are useful for this stress test because they include diverse recurrent, convolutional, and attention-based architectures, while also being designed for efficient sequence processing.

\subsection{Network Changes}

Only minor model changes were required to adapt the speech-to-text architectures to pose forecasting.
The data pipeline was replaced so that motion sequences rather than audio features were loaded, and the task-specific input and output layers were adjusted to predict joint coordinates.
The hidden dimensions and filter shapes were modified to keep layer and model sizes comparable across variants.
All \textit{BatchNorm} layers were removed because they consistently degraded training.
For \textit{Conformer} and \textit{Squeezeformer}, the initial subsampling layers were changed from $4\times$ to $2\times$ temporal downsampling to match the required output sequence length.
As with the other models in the unified benchmark, all \textit{Human3.6M} models were trained from scratch.

\subsection{Ablation Studies}
\label{app:ablations}

As can be seen in Table~\ref{tab:abs_stt}, most converted speech models are very fast, which leaves room to increase the model size further.
With that in mind, the \textit{Conformer} model was improved by increasing the layer dimensions.
As another change, because pose sequences are notably shorter than typical audio sequences, the time reduction at the start of the model was moved to the end, so that temporal information is retained longer in the network.
For the same reason, larger speech models with stronger temporal compression were not considered here.
The resulting model is called \textit{MotionConformer}.
Although it is the second-largest model overall, its training time of only $1h$ on the Nvidia-RTX4080 is on the lower end among the compared architectures.
As shown in Table~\ref{tab:impabs_stt}, it obtains the lowest reported error among the tested models while maintaining high inference speed.

\begin{table}[H]
   \centering
   \fontsize{7.3pt}{9pt}\selectfont
   \begin{tabular}{@{}|l|c|c|r|c|@{}}
      \toprule
      Method                              & Size  & MPJPE & FPS  & \hspace{-1pt}FADE\hspace{-1pt} \\
      \midrule
      Conformer                           & 3.22M &  147         & 1049 &   147 \\
      { }increased model size             & 8.41M &  144         & 932  &  144  \\
      { }moved time reduction             & 3.59M &  146         & 965  &  146  \\
      \midrule
      MotionConformer                     & 9.23M & \textbf{143} & 929  & \textbf{143} \\
      { }no spec augmentation             &       & 144   &       &      \\
      { }no convolutional modules         & 7.89M & 145   &  1024 &  145 \\
      { }xxl model size                   & 122M  & 145   &  264  &  146 \\
      \midrule
      5-run averages &&&& \\
      { }EMPMP                            &       & $152.3 \pm 0.94$ &  & 153.0 \\
      { }EqMotion                         &       & $148.1 \pm 1.19$ &  & 151.4 \\
      { }Conformer (default)              &       & $147.6 \pm 0.24$ &  & 147.7 \\
      { }Conformer (increased size)       &       & $145.4 \pm 0.47$ &  & 145.6 \\
      { }MotionConformer                  &       & $144.0 \pm 0.45$ &  & 144.2 \\
      \bottomrule
   \end{tabular}
   \vspace{6pt}
   \caption{Implemented improvements and ablations of \textit{MotionConformer}, tested on \textit{Human3.6M}.}
   \label{tab:impabs_stt}
\end{table}

The ablations in Table~\ref{tab:impabs_stt} are primarily intended as exploratory checks, since some observed differences are small.
The transition from the adapted \textit{Conformer} to \textit{MotionConformer} gives the clearest improvement, while the individual effects of the later ablations are less conclusive.
\textit{SpecAug}-style masking is a plausible augmentation for pose forecasting and should be investigated more closely in future work.
It randomly masks contiguous parts of the input sequence along the temporal or channel axis, encouraging the model to handle missing timesteps or missing joints.
The role of the \textit{Conformer}-specific convolutional modules was also evaluated.
Removing the convolutional modules from the \textit{Conformer} blocks only slightly changes the result in this measurement.
While this leaves open whether they provide a small benefit, it indicates that the convolutional components are not the dominant source of the model's performance.
Further increasing the model size did not improve the performance, but only reduces the inference speed, which suggests that the model is already sufficiently large for this task.
A five-seed rerun of the final \textit{MotionConformer} configuration shows stable training.
The gap between the mean and best checkpoint is small relative to the gap to the next-best evaluated models, and the best checkpoint is released for direct artifact-level reproducibility.
The multi-seed-runs of the other models showed similar stability, and further strengthen the comparatively better performance of \textit{MotionConformer}.


\vspace{-2pt}
\section{Artificial Noise Pretraining}
\label{app:noise_pretraining}
\vspace{-1pt}

As a lightweight robustness baseline, this appendix evaluates whether pretraining with synthetic coordinate noise improves performance under detector-generated inputs.
Even though adding artificial Gaussian noise to the joint coordinates does not match the errors from a pose estimation system, it still can be used to prepare the forecasting network for such type of input.
Finetuning the networks with random noise (\textit{std:}~$25mm$, clip at~$125mm$) notably improves the zero-shot performance for longer~($1000ms$) predictions, as can be seen in Table~\ref{tab:zsn_10}. 
The training of \textit{EqMotion} always diverged mid-training, so no results are available for this model here.
As a first side note, further experiments with the other two models showed that in both cases, the performance on the clean ground-truth dataset slightly dropped after adding noise, but this is expectable because of the then lower train-test similarity.
As a second side note, adding random position noise directly to the joint positions is preferable to using limb-rotation-based noise, because while axis noise mathematically also just changes the joint position in the end, it keeps the limb lengths correct, which is not true for detector noise, and therefore not desired.

\begin{table}[H]
   \centering
   \fontsize{8pt}{9pt}\selectfont
   \begin{tabular}{|l|cc|cc|}
      \toprule
      Method & \multicolumn{2}{c|}{detector target} & \multicolumn{2}{c|}{ground-truth target} \\
             & { } $400ms$ & $1000ms$ & { }{ }$400ms$ & $1000ms$ \\
      \midrule
      EMPMP                                    & { } 82.5 & 188            & { }{ }{ } 108 & 204 \\
      MotionConformer                          & { } 82.4 & 185            & { }{ }{ } 106 & 199 \\
      \bottomrule
   \end{tabular}
   \vspace{6pt}
   \caption{Improvements through artificial noise pre-training. The left/right columns are separated into prediction/groundtruth targets, as in Table~\ref{tab:zs_10}, showing the measurable/real error.}
   \label{tab:zsn_10}
\end{table}


\vspace{-15pt}
\section{Impact of Joint Errors on Forecasting}

\looseness=-1
Regarding the influence of the noise, the influence of missing joints should be neglectable (less than $0.2\%$ of the joints have an error above $500mm$, which is usually counted as "missing"). 
As only around $12\%$ of the input joints have a displacement error above $10cm$, the overall impact of very poor joints should not the main extra MPJPE reason (as the experiments in the \textit{RapidPoseTriangulation}~\cite{rapidtriang} showed this level is also expectable (or even better) for other datasets). 
The additional error caused by the noisy joints is also evenly distributed over most joints types (with wrist and nose a bit higher).
Regarding the actions, the ones with a notably higher joint displacement (above $10cm$, greeting, sittingdown, waiting) have a slightly higher average difference, but it seems in general that the errors were well tolerated as the model can use a full sequence of $50$ input frames.
Most impact on the MPJPE is caused by the action complexity itself.

\begin{table}[H]
   \centering
   {
   \setlength{\tabcolsep}{4pt}
   \fontsize{8}{8}\selectfont
\begin{tabular}{@{}|l|c|c|c|c|c|@{}}
   \toprule
   Action-Class  & MPJPE   & MPJPE   & above & MPJPE   & Difference \\
   Joint-Type    & (clean) & (input) & 10cm \% & (noisy) & (noisy-clean) \\
   \midrule
   Directions    & 140 & 54  & 5  & 156 & 16 \\
   Discussion    & 174 & 53  & 5  & 196 & 22 \\
   Eating        & 90  & 52  & 6  & 103 & 13 \\
   Greeting      & 235 & 96  & 37 & 262 & 27 \\
   Phoning       & 101 & 57  & 8  & 124 & 23 \\
   Posing        & 174 & 51  & 6  & 194 & 20 \\
   Purchases     & 179 & 56  & 8  & 203 & 24 \\
   Sitting       & 63  & 75  & 15 & 93  & 30 \\
   SittingDown   & 99  & 112 & 43 & 120 & 21 \\
   Smoking       & 106 & 55  & 9  & 125 & 19 \\
   TakingPhoto   & 227 & 56  & 6  & 244 & 17 \\
   Waiting       & 164 & 96  & 35 & 196 & 32 \\
   WalkDog       & 147 & 48  & 4  & 170 & 23 \\
   Walking       & 225 & 52  & 4  & 255 & 30 \\
   WalkTogether  & 102 & 49  & 6  & 116 & 14 \\
   Average       & 143 & 63  & 12 & 165 & 22 \\
   \midrule
   hip\_right      & 104 & 91  & 18 & 125 & 21 \\
   hip\_left       & 103 & 70  & 11 & 124 & 21 \\
   knee\_right     & 111 & 48  & 7  & 131 & 20 \\
   knee\_left      & 109 & 55  & 7  & 126 & 17 \\
   ankle\_right    & 112 & 81  & 10 & 128 & 16 \\
   ankle\_left     & 114 & 97  & 13 & 124 & 10 \\
   nose            & 146 & 105 & 42 & 189 & 43 \\
   shoulder\_right & 138 & 47  & 8  & 157 & 19 \\
   shoulder\_left  & 136 & 43  & 7  & 152 & 16 \\
   elbow\_right    & 179 & 45  & 8  & 194 & 15 \\
   elbow\_left     & 175 & 47  & 8  & 189 & 14 \\
   wrist\_right    & 221 & 45  & 11 & 248 & 27 \\
   wrist\_left     & 209 & 45  & 9  & 235 & 26 \\
   \bottomrule
\end{tabular}
   }
   \vspace{6pt}
   \caption{Errors of \textit{MotionConformer} per action and joint type on the \textit{Human3.6M} dataset using ground-truth or detector-noisy joints.}
   \label{tab:error_per_action}
\end{table}


\vspace{-15pt}
\section{Example of Pseudo-Label Adaptation}
\label{app:noise_vis}

\begin{figure}[H]
   \centering
   \begin{subfigure}{.37\linewidth}
     \centering
     \includegraphics[width=\linewidth]{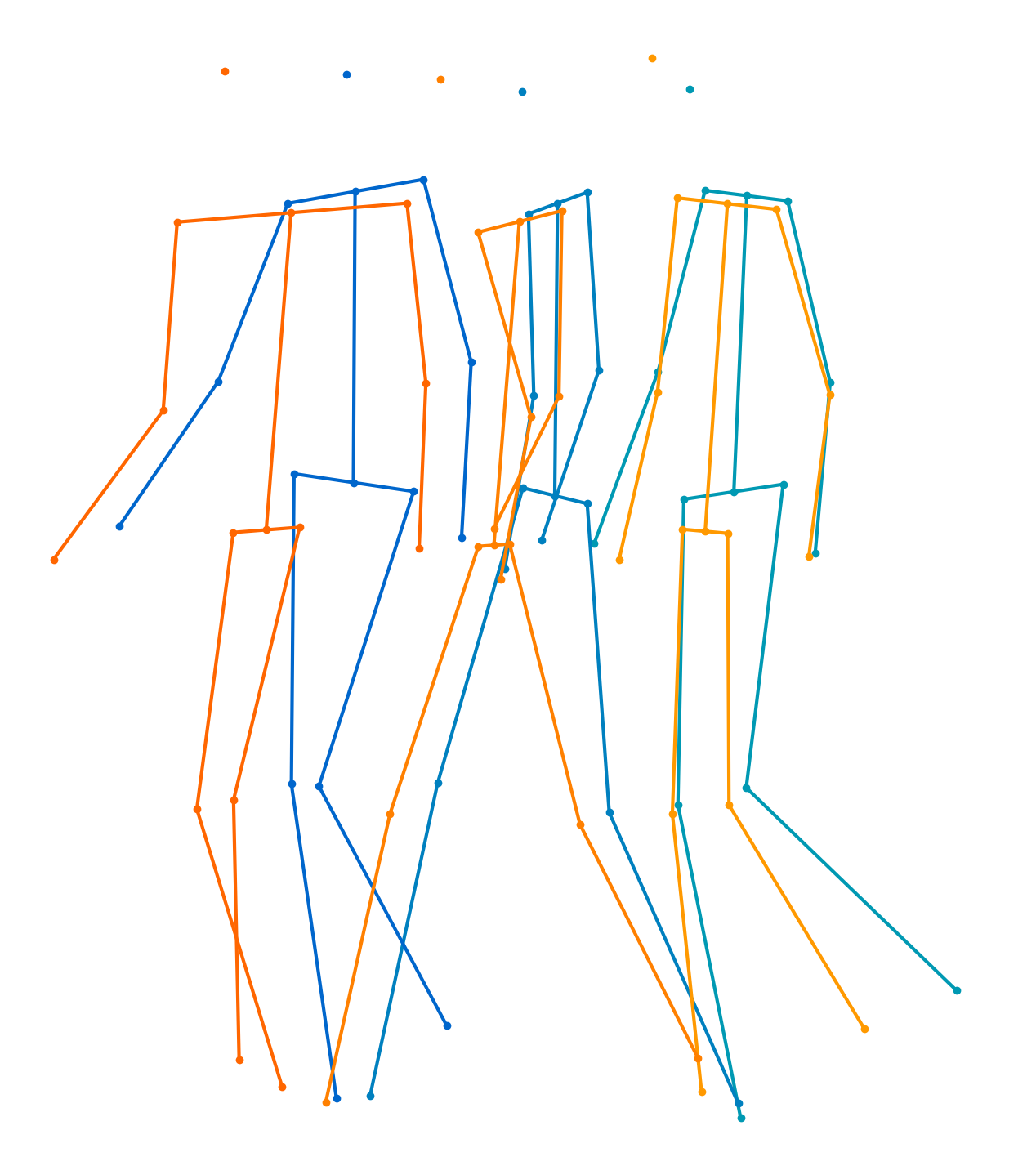}
     \caption{Zero-Shot}
     \label{fig:sub1a}
   \end{subfigure}%
   \begin{subfigure}{.33\linewidth}
     \centering
     \includegraphics[width=\linewidth]{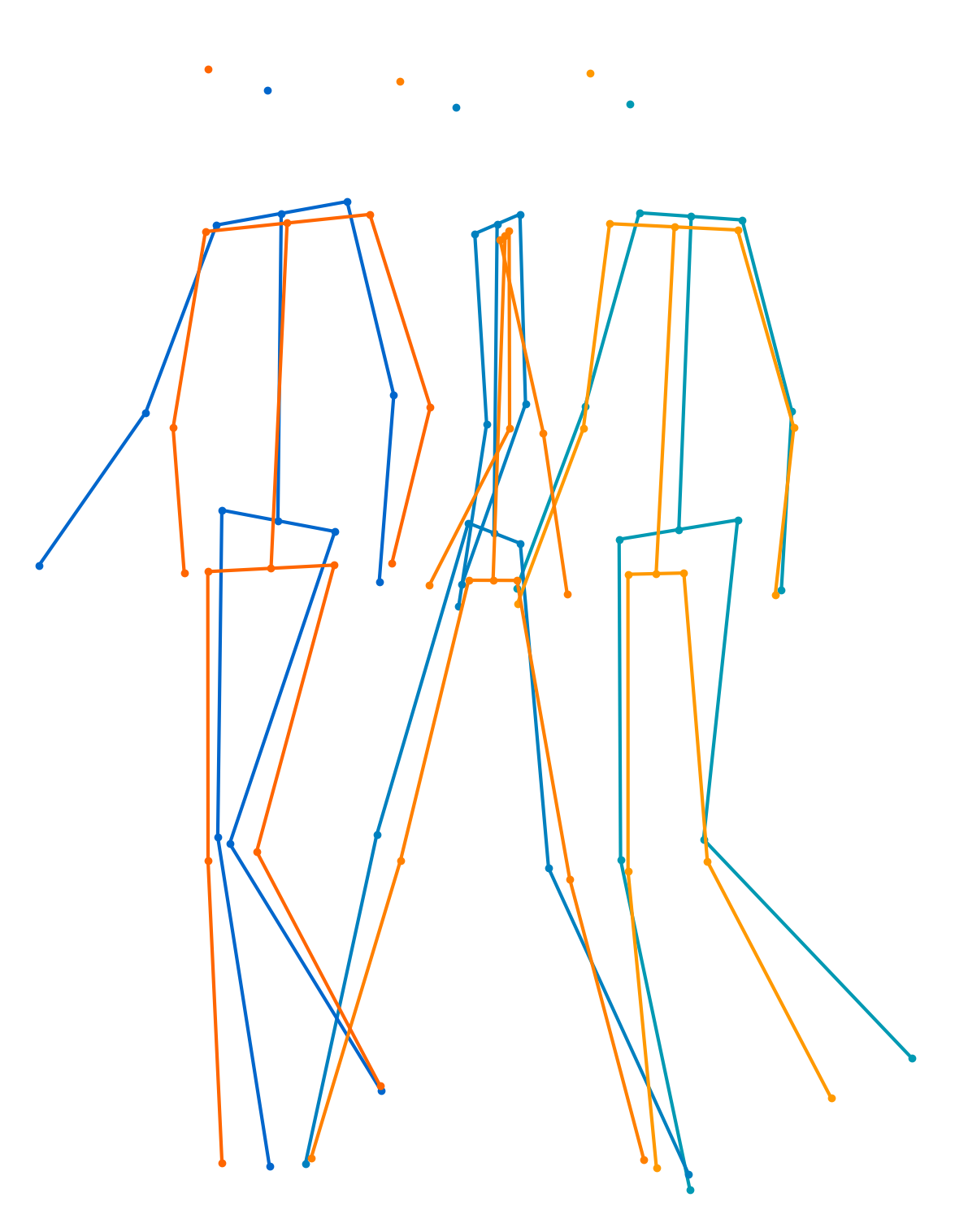}
     \caption{Finetuned}
     \label{fig:sub1b}
   \end{subfigure}
   \vspace{6pt}
   \caption{Example of improvements through finetuning on \textit{Human3.6M} dataset. In blue the ground-truth, in orange the prediction. The walking movement was already well continued in (a), but especially the stride distances improved after finetuning. For better visualization some intermediate timesteps are not displayed.}
   \label{fig:comp1}
\end{figure}

Figure~\ref{fig:comp1} shows one representative forecast before and after adaptation on detector-generated pose sequences.
While this example is only illustrative, it visualizes the same effect observed quantitatively in Section~\ref{chap:noisy}.

\end{document}